\documentclass[twoside,11pt]{article}

\usepackage{dnd}
\usepackage{times}
\usepackage{linguex}
\usepackage{xcolor}
\usepackage{mathtools}
\usepackage{todonotes}
\usepackage{multirow}
\usepackage{amsmath}
\usepackage[bottom]{footmisc}
\usepackage{array}

\def\doubleunderline#1{\underline{\underline{#1}}}

\dndheading{issue(number)}{year}{firstpage--lastpage}{Name1 Surname1, Name2 Surname2, and Name3 Surname3}{10.5087/dad.DOINUMBER}

\ShortHeadings{A Neural Approach to Discourse Relation Signal Detection}{
Zeldes and Liu
}
\firstpageno{1}

\begin{document}

\title{A Neural Approach to Discourse Relation Signal Detection}

\author{\name Amir Zeldes \email amir.zeldes@georgetown.edu \\
       \addr Georgetown University\\
       Department of Linguistics
       \AND
       \name Yang (Janet) Liu \email yl879@georgetown.edu \\
       \addr Georgetown University\\
       Department of Linguistics
       }

\editor{Name Surname}
\submitted{MM/YYYY}{MM/YYYY}{MM/YYYY}

\maketitle

\begin{abstract}%
Previous data-driven work investigating  the types and distributions of discourse relation signals, including discourse markers such as `however' or phrases such as `as a result' has focused on the relative frequencies of signal words within and outside text from each discourse relation. Such approaches do not allow us to quantify the signaling strength of individual instances of a signal on a scale (e.g.~more or less discourse-relevant instances of `and'), to assess the distribution of ambiguity for signals, or to identify words that hinder discourse relation identification in context (`anti-signals' or `distractors'). In this paper we present a data-driven approach to signal detection using a distantly supervised neural network and develop a metric, ${\Delta}_s$ (or `delta-softmax'), to quantify signaling strength. Ranging between -1 and 1 and relying on recent advances in contextualized words embeddings, the metric represents each word's positive or negative contribution to the identifiability of a relation in specific instances in context. Based on an English corpus annotated for discourse relations using Rhetorical Structure Theory and signal type annotations anchored to specific tokens, our analysis examines the reliability of the metric, the places where it overlaps with and differs from human judgments, and the implications for identifying features that neural models may need in order to perform better on automatic discourse relation classification.

\end{abstract}


\section{Introduction} \label{introduction}

The development of formal frameworks for the analysis of discourse relations has long gone hand in hand with work on signaling devices. The analysis of discourse relations is also closely tied to what a discourse structure should look like and what discourse goals should be fulfilled in relation to the interpretation of discourse relations \citep{roberts:2012:information}. Earlier work on the establishment of inventories of discourse relations and their formalization (\citealt{HovyMaier1993}, \citealt{KnottDale1994}, \citealt{Knott1996}, \citealt{knott1998classification}, \citealt{webber1998anchoring}, \citealt{Fraser1999}) relied on the existence of `discourse markers' (DMs) or `connectives', including conjunctions such as \textit{because} or \textit{if}, adverbials such as \textit{however} or \textit{as a result}, and coordinations such as \textit{but}, to identify and distinguish relations such as \textsc{condition} in  \ref{ex:condition},  \textsc{concession} in \ref{ex:concession}, \textsc{cause} in \ref{ex:cause}, or \textsc{contrast}, \textsc{result} etc., depending on the postulated inventory of relations (signals for these relations as identified by human analysts are given in bold; examples come from the GUM corpus \citep{zeldes2017gum}, presented in Section \ref{data}).

\ex. [\textbf{If} you work for a company,]$_{\textsc{condition}}$ [they pay you that money.]\label{ex:condition}

\ex. [\textbf{Albeit} limited,]$_{\textsc{concession}}$ [these results provide valuable insight into SI interpretation by Chitonga-speaking children.] \label{ex:concession}

\ex. [not all would have been interviewed at Wave 3] [\textbf{due to} differential patterns of temporary attrition]$_{\textsc{cause}}$\label{ex:cause}

The same reasoning of identifying relations based on overt signals has been applied to the comparison of discourse relations across languages, by comparing inventories of similar function words cross-linguistically (\citealt{TaboadaAngelesGomez-Gonzalez}, \citealt{VersleyGastel2013}); and the annotation guidelines of prominent contemporary corpora rely on such markers as well: for instance, the Penn Discourse Treebank (see \citealt{PrasadEtAl2008}) explicitly refers to either the presence of DMs or the possibility of their insertion in cases of implicit discourse relations, and DM analysis in Rhetorical Structure Theory \citep{MannThompson1988} has also shown the important role of DMs as signals of discourse relations at all hierarchical levels of discourse analysis \citep{DasTaboada2017}. 

At the same time, research over the past two decades analyzing the full range of possible cues that humans use to identify the presence of discourse relations has suggested that classic DMs such as conjunctions and adverbials are only a part of the network of signals that writers or speakers can harness for discourse structuring, which also includes entity-based cohesion devices (e.g.~certain uses of anaphora, see \citealt{PoesioEugenioKeohane2002}), alternative lexicalizations using content words, as well as syntactic constructions (see \citealt{PrasadWebberJoshi2014} and the addition of alternative lexicalization constructions, AltLexC, in the latest version of PDTB, \citealt{PrasadWebberLeeEtAl2019}).\footnote{In the examples above as well, we can identify non-DM signals such as the concessive word \textit{limited} in the \textsc{concession} relation in \ref{ex:concession}, or even the use of the generic present tense and the generic `you' in ``\textit{if you work}'' in \ref{ex:condition} as markers of general conditional clauses. Finding and agreeing on the exact span of words that constitutes a signal, or `signal anchoring' (see Section \ref{sec:anchored_gum}) are therefore non-trivial tasks.} 

In previous work, two main approaches to extracting the inventory of discourse signal types in an open-ended framework can be identified: data-driven approaches, which attempt to extract relevant words from distributional properties of the data, using frequencies or association measures capturing their co-occurrences with certain relation types (e.g.~\citealt{TorabiAsrDemberg2013}, \citealt{ToldovaPisarevskayaAnanyevaEtAl2017}); and manual annotation efforts (e.g.~\citealt{PrasadEtAl2008}, \citealt{TaboadaDas2013}), which develop categorization schemes and guidelines for human evaluation of signaling devices. The former family of methods benefits from an unbiased openness to any and every type of word which may reliably co-occur with some relation types, whether or not a human might notice it while annotating, as well as the naturally graded and comparable nature of the resulting quantitative scores, but, as we will show, falls short in identifying specific cases of a word being a signal (or not) in context. By contrast, the latter approach allows for the identification of individual instances of signaling devices, but relies on less open-ended guidelines and is categorical in nature: a word either is or isn't a signal in context, providing less access to concepts such as signaling strength. The goal of this paper is to develop and evaluate a model of discourse signal identification that is built bottom up from the data, but retains sensitivity to context in the evaluation of each individual example.\footnote{\cite{zeldes2018neural} showed some preliminary results from the early stages of this line of work.} In addition, even though this work is conducted within Rhetorical Structural Theory, we hope that it can shed light on signal identification of discourse relations across genres and provide empirical evidence to motivate research on theory-neutral and genre-diverse discourse processing, which would be beneficial for pushing forward theories of discourse across frameworks or formalisms. Furthermore, employing a computational approach to studying discourse relations has a promising impact on various NLP downstream tasks such as question answering and document summarization etc. For example, \cite{narasimhan-barzilay-2015-machine} incorporated discourse information into the task of automated text comprehension and benefited from such information without relying on explicit annotations of discourse structure during training, which outperformed state-of-the-art text comprehension systems at the time.

Towards this goal, we begin by reviewing some previous work in the traditions sketched out above in the next section, and point out some open questions which we would like to address. In Section \ref{data} we present the discourse annotated data that we will be using, which covers a number of English text types from the Web annotated for 20 discourse relations in the framework of Rhetorical Structure Theory, and is enriched with human annotations of discourse relation signaling devices for a subset of the data. Moreover, we also propose a taxonomy of anchored signals based on the discourse annotated data used in this paper, illustrating the properties and the distribution of the anchorable signals.

In Section \ref{automatic_signal_extraction} we then train a distantly supervised neural network model which is made aware of the relations present in the data, but attempts to learn which words signal those relations without any exposure to explicit signal annotations. We evaluate the accuracy of our model using state-of-the-art pretrained and contextualized character and word embeddings, and develop a metric for signaling strength based on a masking concept similar to permutation importance, which naturally lends itself to the definition of both positive and negative or `anti-signals', which we will refer to as `distractors'. 

In Section \ref{evaluation}, we combine the anchoring annotation data from Section \ref{data} with the model's predictions to evaluate how `human-like' its performance is, using an information retrieval approach measuring recall@k and assessing the stability of different signal types based on how the model scores them. We develop a visualization for tokenwise signaling strength and perform error analysis for some signals found by the model which were not flagged by humans and vice versa, and point out the strengths and weaknesses of the architecture. Section \ref{discussion} offers further discussion of what we can learn from the model, what kinds of additional features it might benefit from given the error analysis, and what the distributions of scores for individual signals can teach us about the ambiguity and reliability of different signal types, opening up avenues for further research.

\section{Previous Work} \label{previous_work}

\subsection{Data-driven Approaches}\label{sec:datadriven}

A straightforward approach to identifying discourse relation signals in corpora with discourse parses is to extract frequency counts for all lexical types or lemmas and cross-tabulate them with discourse relations, such as sentences annotated as \textsc{cause}, \textsc{elaboration}, etc. (e.g.~\citealt{SchefflerStede2016}, \citealt[73]{MirovskySynkovaRysovaEtAl2017}, \citealt{ToldovaPisarevskayaAnanyevaEtAl2017}). Table \ref{tab:russian}, reproduced from \citet[32]{ToldovaPisarevskayaAnanyevaEtAl2017}, illustrates this approach for the Russian RST Treebank.

\begin{table}[htb]
\centering
\begin{tabular}{l|r|l|l}
relation type & freq & marker & translation \\
\hline
Elaboration & 150 & \textit{kotoryj} & which, that \\
Joint & 119 & \textit{i, takzhe} & and, as well \\
Attributon & 118 & \textit{zajavil, soobschil} & report, announce etc. \\
Contrast & 61 & \textit{Odnako, a, no} & However, but \\
Cause-Effect & 47 & \textit{Poetomu, V+prichina} & so, accordingly, V+cause \\
Purpose & 39 & \textit{Chtoby, dlya} & In order that, for \\
Interpretation-Evaluation & 34 & Nouns / verbs expressing opinion &  \\
Background & 31 & No dominant marker &  \\
Condition & 27 & \textit{esli} & if \\
\end{tabular}
\caption{Russian discourse relation signals, reproduced from \citet{ToldovaPisarevskayaAnanyevaEtAl2017}.}
\label{tab:russian}
\end{table}

This approach quickly reveals the core inventory of cue words in the language, and in particular the class of low-ambiguity discourse markers (DMs), such as \textit{odnako} `however' signaling \textsc{contrast} (see Fraser 1999 on delimiting the class of explicit DMs) or relative pronouns signaling \textsc{elaboration}. As such, it can be very helpful for corpus-based lexicography of discourse markers (cf.~\citealt{StedeUmbach1998}). The approach can potentially include multiword expressions, if applied equally to multi-token spans (e.g.~\textit{as a result}), and because it is language-independent, it also allows for a straightforward comparison of connectives or other DMs across languages. Results may also converge across frameworks, as the frequency analysis may reveal the same items in different corpora annotated using different frameworks. For instance, the inventory of connectives found in work on the Penn Discourse Treebank (PDTB, see \citealt{PrasadEtAl2008}) largely converges with findings on connectives using RST (see \citealt[97--109]{Stede2012}, \citealt{TaboadaDas2013}): conjunctions such as \textit{but} can mark different kinds of contrastive relations at a high level, and adverbs such as \textit{meanwhile} can convey contemporaneousness, among other things, even when more fine-grained analyses are applied. However, a purely frequentist approach runs into problems on multiple levels, as we will show in Section \ref{automatic_signal_extraction}: high frequency and specificity to a small number of relations characterize only the most common and unambiguous discourse markers, but not less common ones. Additionally, differentiating actual and potentially ambiguous usages of candidate words in context requires substantial qualitative analysis (see \citealt{Cunha2013}), which is not reflected in aggregated counts, and signals that belong to a class of relations (e.g.~a variety of distinct contrastive relation types) may appear to be non-specific, when in fact they reliably mark a superset of relations.

Other studies have used more sophisticated metrics, such as point-wise mutual information (PMI), to identify words associated with particular relations \citep{TorabiAsrDemberg2013}. Using the PDTB corpus, \citet{TorabiAsrDemberg2013} extracted such scores and measured the contribution of different signal types based on the information gain which they deliver for the classification of discourse relations at various degrees of granularity, as expressed by the hierarchical labels of PDTB relation types. This approach is most similar to the goal given to our own model in Section \ref{automatic_signal_extraction}, but is less detailed in that the aggregation process assigns a single number to each candidate lexical item, rather than assigning contextual scores to each instance.\footnote{In this overview we disregard supervised approaches to detecting signals in the sense of connective detection, which rely primarily on labeled training data. Such approaches can be very effective in reproducing benchmark annotations from corpora such as PDTB, but do so by effectively memorizing lexical items from gold standard annotated data (see \citealt{MullerBraudMorey2019} for recent state-of-the art work). As such, they are less interesting for open-ended exploration of signaling devices.} 

Finally we note that for hierarchical discourse annotation schemes, the data-driven approaches described here become less feasible at higher levels of abstraction, as relations connecting entire paragraphs encompass large amounts of text, and it is therefore difficult to find words with high specificity to those relations. As a result, approaches using human annotation of discourse relation signals may ultimately be irreplaceable.

\subsection{Discourse Relation Signal Annotations}

Discourse relation signals are broadly classified into two categorizes: \textit{anchored} signals and \textit{unanchored} signals. By `anchoring' we refer to associating signals with concrete token spans in texts. Intuitively, most of the signals are anchorable since they correspond to certain token spans. However, it is also possible for a discourse relation to be signaled but remain unanchored. Results from \cite{liu-2019-beyond} indicated that there are several signaled but unanchored relations such as \textsc{preparation} and \textsc{background} since they are high-level discourse relations that capture and correspond to genre features such as \textit{interview layout} in interviews where the conversation is constructed as a question-answer scheme, and are thus rarely anchored to tokens. 

The Penn Discourse Treebank (PDTB V3, \citealt{PrasadWebberLeeEtAl2019}) is the largest discourse annotated corpus of English, and the largest resource annotated explicitly for discourse relation signals such as connectives, with similar corpora having been developed for a variety of languages (e.g.~\citealt{ZeyrekDemirsahinSevdik-CalliEtAl2013} for Turkish,  \citealt{ZhouLuZhangEtAl2014} for Chinese). However the annotation scheme used by PDTB is ahierarchical, annotating only pairs of textual argument spans connected by a discourse relation, and disregarding relations at higher levels, such as relations between paragraphs or other groups of discourse units. Additionally, the annotation scheme used for explicit signals is limited to specific sets of expressions and constructions, and does not include some types of potential signals, such as the graphical layout of a document, lexical chains of (non-coreferring) content words that are not seen as connectives, or genre conventions which may signal the discourse function for parts of a text. It is nevertheless a very useful resource for obtaining frequency lists of the most prevalent DMs in English, as well as data on a range of phenomena such as anaphoric relations signaled by entities, and some explicitly annotated syntactic constructions. 

Working in the hierarchical framework of Rhetorical Structure Theory \citep{MannThompson1988}, \citet{TaboadaDas2013} re-annotated the existing RST Discourse Treebank \citep{CarlsonEtAl2003}, by taking the existing discourse relation annotations in the corpus as a ground truth and analyzing any possible information in the data, including content words, patterns of repetition or genre conventions, as a possibly present discourse relation signaling device. The resulting RST Signalling Corpus (RST-SC, \citealt{DasTaboadaMcFetridge2019}) consists of 385 Wall Street Journal articles from the Penn Treebank \citep{MarcusSantoriniMarcinkiewicz1993}, a smaller subset of the same corpus used in PDTB. It contains 20,123 instances of 78 relation types (e.g.~\textsc{attribution}, \textsc{circumstance}, \textsc{result} etc.), which are enriched with 29,297 signal annotations. \citet{DasTaboada2017} showed that when all types of signals are considered, over 86\% of discourse relations annotated in the corpus were signaled in some way, but among these, just under 20\% of cases were marked by a DM. However, unlike PDTB, the RST Signalling Corpus does not provide a concrete span of tokens for the locus of each signal, indicating instead only the type of signaling device used.

Although the signal annotations in RST-SC have a broader scope than those in PDTB and are made more complex by extending to hierarchical relations, \citet{liu2019discourse} have shown that RST-SC's annotation scheme can be `anchored' by associating discourse signal categories from RST-SC with concrete token spans. \citet{liu-2019-beyond} applied the same scheme to a data set described in Section \ref{data}, which we will use to evaluate our model in Section \ref{evaluation}. Since that data set is based on the same annotation scheme of signal types as RST-SC, we will describe the data for the present study and RST-SC signal type annotation scheme next.

\section{Data} \label{data}


\subsection{Anchored Signals in the GUM Corpus}\label{sec:anchored_gum}

In order to study open-ended signals anchored to concrete tokens, we use the signal-annotated subset of the freely available Georgetown University Multilayer (GUM) Corpus \citep{zeldes2017gum} from \citet{liu-2019-beyond}. Our choice to use a multi-genre RST-annotated corpus rather than using PDTB, which also contains discourse relation signal annotation to a large extent is motivated by three reasons: The first reason is that we wish to explore the full range of potential signals, as laid out in the work on the Signalling Corpus \citep{DasTaboada2017, das2018rst}, whereas PDTB annotates only a subset of the possible cues identified by human annotators. Secondly, the use of RST as a framework allows us to examine discourse relations at all hierarchical levels, including long distance, high-level relations between structures as large as paragraphs or sections, which often have different types of signals allowing their identification. Finally, although the entire GUM corpus is only about half the size of RST-DT (109K tokens\footnote{During the review period, the corpus has since grown to 130K tokens, however we were not able to include these in the analyses presented here.}), using GUM offers the advantage of a more varied range of genres than PDTB and RST-SC, both of which annotate Wall Street Journal data. 

The signal annotated subset of GUM includes academic papers, how-to guides, interviews and news text, encompassing over 11,000 tokens. Although this data set may be too small to train a successful neural model for signal detection, we will not be using it for this purpose; instead, we will reserve it for use solely as a test set, and use the remainder of the data (about 98K tokens) to build our model (see Section \ref{neuralmodel} for more details about the subsets and splits), including data from four further genres, for which the corpus also contains RST annotations but no signaling annotations: travel guides, biographies, fiction, and Reddit forum discussions. The GUM corpus is manually annotated with a large number of layers, including document layout (headings, paragraphs, figures, etc.); multiple POS tags (Penn tags, CLAWS5, Universal POS); lemmas; sentence types (e.g.~imperative, wh-question etc., \citealt{LeechEtAl2003}); Universal Dependencies \citep{NivreEtAl2017}; (non-)named entity types; coreference and bridging resolution; and discourse parses using Rhetorical Structure Theory \citep{MannThompson1988}. In particular, the RST annotations in the corpus use a set of 20 commonly used RST relation labels, which are given in Table \ref{tab:reltypes}, along with their frequencies in the corpus.\footnote{During the review period, the RST component has undergone segmentation and relation set changes (there are now 25 labels in the corpus), however we were not able to include these in the analyses presented here.} The relations cover asymmetrical prominence relations (satellite-nucleus) and symmetrical ones (multinuclear relations), with the \textsc{restatement} relation being realized in two versions, one for each type.

\begin{table}[h!tb]
\centering
\begin{tabular}{l|c|c|l|l|c}
relation & type & \# & relation & type & \# \\ \hline
\textsc{joint} & multinuclear & 1248 & \textsc{condition} & satellite-nucleus & 210 \\
\textsc{elaboration} & satellite-nucleus & 1037 & \textsc{justify} & satellite-nucleus & 203 \\
\textsc{sequence} & multinuclear & 546 & \textsc{result} & satellite-nucleus & 185 \\
\textsc{preparation} & satellite-nucleus & 500 & \textsc{solutionhood} & satellite-nucleus & 173 \\
\textsc{background} & satellite-nucleus & 464 & \textsc{restatement} & multinuclear or & 150 \\
\textsc{circumstance} & satellite-nucleus & 314 &  & satellite-nucleus &  \\
\textsc{evaluation} & satellite-nucleus & 283 & \textsc{evidence} & satellite-nucleus & 147 \\
\textsc{concession} & satellite-nucleus & 271 & \textsc{purpose} & satellite-nucleus & 136 \\
\textsc{contrast} & multinuclear & 250 & \textsc{motivation} & satellite-nucleus & 124 \\
\textsc{cause} & satellite-nucleus & 242 & \textsc{antithesis} & satellite-nucleus & 109 \\
\end{tabular}
\caption{RST relations and their frequencies in the GUM corpus.}
\label{tab:reltypes}
\end{table}




The signaling annotation in the corpus follows the scheme developed by RST-SC, with some additions. Although RST-SC does not indicate token positions for signals, it provides a detailed taxonomy of signal types which is hierarchically structured into three levels: \\

\begin{enumerate}
    \item  \textit{signal class}, denoting the signal's degree of complexity
    \item \textit{signal type}, indicating the linguistic system to which it belongs
    \item \textit{specific signal}, which gives the most fine-grained subtypes of signals within each type
\end{enumerate}

It is assumed that any number of word tokens can be associated with any number of signals (including the same tokens participating in multiple signals), that signals can arise without corresponding to specific tokens (e.g.~due to graphical layout of paragraphs), and that each relation can have an unbounded number of signals ($0-n$), each of which is characterized by all three levels.

The \textit{signal class} level is divided into \textit{single}, \textit{combined} (for complex signals), and \textit{unsure} for unclear signals which cannot be identified conclusively, but are noted for further study. For each signal (regardless of its class), \textit{signal type} and \textit{specific signal} are identified. According to RST-SC's taxonomy, \textit{signal type} includes 9 types such as \textit{DMs, genre, graphical, lexical, morphological, numerical, reference, semantic,} and \textit{syntactic}. Each type then has specific subcategories. For instance, the signal type \textit{semantic} has 7 specific signal subtypes: \textit{synonymy, antonymy, meronymy, repetition, indicative word pair, lexical chain}, and \textit{general word}. We will describe some of these in more depth below.

In addition to the 9 signal types, RST-SC has 6 combined signal types such as \textit{reference+syntactic}, \textit{semantic+syntactic}, and \textit{graphical+syntactic} etc., and 15 specific signals are identified for the \textit{combined} signals. Although the rich signaling annotations in RST-SC offer an excellent overview of the relative prevalence of different signal types in the Wall Street Journal corpus, it is difficult to apply the original scheme to the study of individual signal words, since actual signal positions are not identified. While recovering these positions may be possible for some categories using the original guidelines,\footnote{Especially DMs, of which 201 lexical types are identified in RST-SC, could be searched for in the relevant text spans. However even for these we would not always be sure which word is intended in spans containing multiple instances of the relevant word.} most signaling annotations (e.g.~lexical chains, repetition) cannot be automatically paired with actual tokens, meaning that, in order to use the original RST-SC for our study, we would need to re-annotate it for signal token positions. As this effort is beyond the scope of our study, we will use the smaller data set with anchored signaling annotations from \citet{liu-2019-beyond}: This data is annotated with the same signal categories as RST-SC, but also includes exact token positions for each signal, including possibly no tokens for unanchorable signals such as some types of genre conventions or graphical layout which are not expressible in terms of specific words.

In order to get a better sense of how the annotations work, we consider example \ref{ex:discrimination}.


\ex. [5] \textit{\textbf{Sociologists have explored}} the adverse consequences of discrimination; [6] \textit{\textbf{psychologists have examined}} the mental processes that underpin conscious and unconscious biases; [7] \textit{\textbf{neuroscientists have examined}} the neurobiological underpinnings of discrimination; [8] and \textit{\textbf{evolutionary theorists have explored}} the various ways that in-group/out-group biases emerged across the history of our species. -- \textsc{joint} [GUM\_academic\_discrimination] \label{ex:discrimination}

\begin{figure}[htb]
    \centering
    \includegraphics[width=110mm]{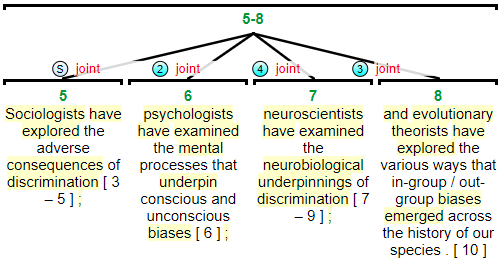}
    \caption{A visualization of an RST analysis of \ref{ex:discrimination} with the signal tokens highlighted.}
    \label{fig:discrimination}
\end{figure}
In this example, there is a \textsc{joint} relation between four spans in a fragment from an RST discourse tree. The first tokens in each span form a parallel construction and include semantically related items such as \textit{explored} and \textit{examined} (signal class `combined', type `semantic+syntactic', specific subtype `parallel syntactic construction + lexical chain'). The words corresponding to this signal in each span are highlighted in Figure \ref{fig:discrimination}, and are considered to signal each instance of the \textsc{joint} relation. Additionally, the \textsc{joint} relation is also signaled by a number of further signals which are highlighted in the figure as well, such as the semicolons between spans, which correspond to a type `graphical', subtype `semicolon' in RST-SC. The data model of the corpus records which tokens are associated with which categorized signals, and allows for multiple membership of the same token in several signal annotations.

In terms of annotation reliability, \citet{DasTaboada2017} reported a weighted kappa of 0.71 for signal subtypes in RST-SC without regard to the span of words corresponding to a signal, while a study by \citet{gessler-etal-2019-discourse}  suggests that signal anchoring, i.e.~associating RST-SC signal categories with specific tokens achieves a 90.9\% perfect agreement score on which tokens constitute signals, or a Cohen's Kappa value of 0.77. As anchored signal positions will be of the greatest interest to our study, we will consider how signal token positions are distributed in the corpus next, and develop an anchoring taxonomy which we will refer back to for the remainder of this paper. 

\subsection{A Taxonomy of Anchored Signals} \label{signal_taxonomy}

From a structural point of view, one of the most fundamental distinctions with regard to signal realization recognized in previous work is the classification of signaling tokens into satellite or nucleus-oriented positions, i.e.~whether a signal for the relation appears within the modifier span or the span being modified \citep{taboada2006discourse}. While some relation types exhibit a strong preference for signal position (e.g.~using a discourse marker such as \textit{because} in the satellite for \textsc{cause}, \citealt{Duque2014}), others, such as \textsc{concession} are more balanced (almost evenly split signals between satellite and nucleus in \citealt{taboada2006discourse}). In this study we would like to further refine the taxonomy of signal positions, breaking it down into several features.

At the highest level, we have the distinction between anchorable and non-anchorable signals, i.e.~signals which correspond to no token in the text (e.g.~genre conventions, graphical layout). Below this level, we follow \citet{taboada2006discourse} in classifying signals as satellite or nucleus-oriented, based on whether they appear in the more prominent Elementary Discourse Unit (EDU) of a relation or its dependent. However, several further distinctions may be drawn:

\begin{itemize}
    \item Whether the signal appears before or after the relation in text order; since we consider the relation to be instantiated as soon as its second argument in the text appears, `before' is interpreted as any token before the second head unit in the discourse tree begins, and `after' is any subsequent token
    \item Whether the signal appears in the head unit of the satellite/nucleus, or in a dependent of that unit; this distinction only matters for satellite or nucleus subtrees that consist of more than one unit
    \item Whether the signal is anywhere within the structure dominated by the units participating in the relation, or completely outside of this structure
\end{itemize}

\begin{table}[b]
\begin{tabular}{|l|l|c|c|c|c|c|c|c|c|c|l|}
\hline
\multirow{5}{*}{} & \multicolumn{10}{l|}{+A(nchorable)} & -A \\ \cline{2-12} 
 & \multicolumn{5}{l|}{Before} & \multicolumn{5}{l|}{After} & \multirow{4}{*}{} \\ \cline{2-11}
 & \multicolumn{4}{l|}{I(nside)} & \multicolumn{1}{l|}{O(utside)} & \multicolumn{4}{l|}{I(nside)} & \multicolumn{1}{l|}{O(utside)} &  \\ \cline{2-11}
 & \multicolumn{2}{l|}{H(ead)} & \multicolumn{2}{c|}{D(ependent)} &  & \multicolumn{2}{l|}{H(ead)} & \multicolumn{2}{c|}{D(ependent)} & \multicolumn{1}{l|}{} &  \\ \cline{2-11}
 & \multicolumn{1}{c|}{Sat} & Nuc & Sat & Nuc &  & Sat & Nuc & Sat & Nuc &  &  \\ \hline
\multicolumn{1}{|c|}{group} & \multicolumn{1}{c|}{I} & II & III & IV & V & VI & VII & VIII & IX & X & XI \\ \hline
\multicolumn{1}{|c|}{\#tok} & \multicolumn{1}{c|}{513} & 1073 & 311 & 648 & 0 & 1656 & 302 & 948 & 1114 & 0 & 0 \\ \hline
\multicolumn{1}{|c|}{\#edu} & \multicolumn{1}{c|}{157} & 281 & 25 & 55 & 0 & 457 & 89 & 109 & 144 & 0 & 57 \\ \hline
\end{tabular}
\caption{An overview of the taxonomy and the distribution of the corresponding anchorable signals attested in the subset of the GUM corpus from \citet{liu-2019-beyond}.}
\label{tab:taxonomy}
\end{table}

\begin{figure}[htb]
\centering
\includegraphics[width=100mm]{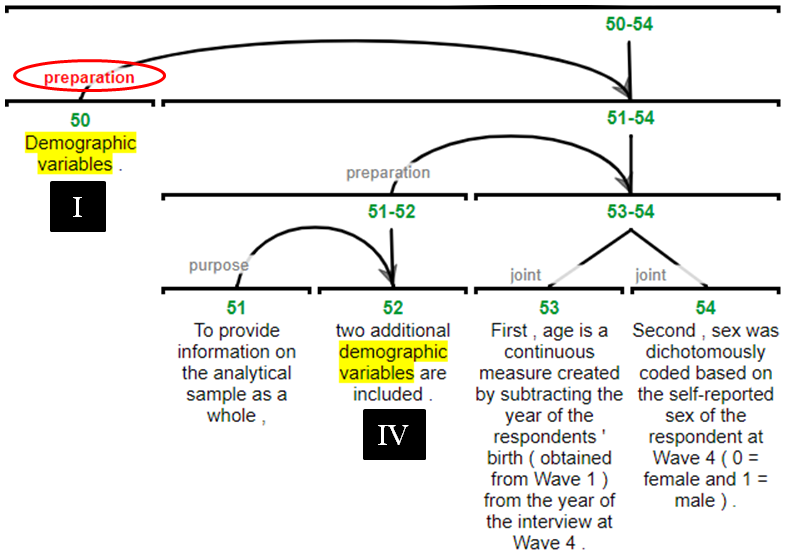}
\caption{Example of a manually annotated signal with token positions corresponding to categories I and IV in the taxonomy with respect to the \textsc{preparation} relation of unit 50.}
\label{fig:taxonomy_example}
\par\vspace{-10pt}\par
\end{figure}

Table \ref{tab:taxonomy} gives an overview of the taxonomy proposed here, which includes the possible combinations of these properties and the distribution of the corresponding anchorable signals found in the signal-annotated subset of the GUM Corpus from \citet{liu-2019-beyond}.\footnote{In this paper we disregard a further set of distinctions pertaining to the position of signals within a discourse unit (unit initial, final, medial etc.), though these too have been explored in the past, especially in the context of cohesive discourse generation \citep{PowerDoranScott1999}.} Individual feature combinations can be referred to either as acronyms, e.g.~ABIHS for `Anchorable, Before the second EDU of the relation, Inside the relation's subtree, Head unit of the Satellite', or using the group IDs near the bottom of the table (in this case the category numbered Roman I). We will refer back to these categories in our comparison of manually annotated and automatically predicted signals. 
To illustrate how the taxonomy works in practice, we can consider the example in Figure \ref{fig:taxonomy_example}, which shows a signal whose associated tokens instantiate categories I and IV in a discourse tree -- the words \textit{demographic variables} appear both within a \textsc{preparation} satellite (unit [50], category I), which precedes and points to its nucleus [51--54], and within a satellite inside that block (unit [52], a dependent inside the nucleus block, category IV). Based on the RST-SC annotation scheme, the signal class is \textit{Simple}, with the type \textit{Semantic} and specific sub-type \textit{Lexical chain}.

The numbers at the bottom of Table \ref{tab:taxonomy} show the number of tokens signaling each relation at each position, as well as the number of relations which have signal tokens at the relevant positions. The hypothetical categories V and X, with signal tokens which are not within the subtree of satellite or nucleus descendants, are not attested in our data, as far as annotators were able to identify.\footnote{A conceivable example would be foreshadowing or summarizing the contents of a relation before or after any of the units involved, such as a discussion of an academic article's sections acting as an early signal of an upcoming heading, which is itself analyzed as a \textsc{preparation}. Though these cases were not found in the small sample used here, they have been studied in the past as a form of metalinguistic labeling (see \cite{Francis1994} for example).}

\section{Automatic Signal Extraction} \label{automatic_signal_extraction}

\subsection{A Contextless Frequentist Approach}

\begin{figure}[t]
\centering
\includegraphics[width=70mm]{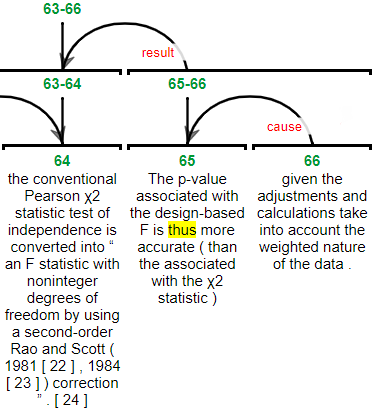}
\caption{RST fragment with signal obeying strong nuclearity: the DM \textit{thus} indicating \textsc{result} is in the head EDU of the satellite block.}
\label{fig:strong_nunclearity}
\par\vspace{-15pt}\par
\end{figure}

To motivate the need for a fine-grained and contextualized approach to describing discourse relation signals in our data, we begin by extracting some basic data-driven descriptions of our data along the lines presented in Section \ref{sec:datadriven}. In order to constrain candidate words to just the most relevant ones for marking a specific signal, we first need a way to address a caveat of the frequentist approach: higher order relations which often connect entire paragraphs (notably \textsc{background} and \textsc{elaboration}) must be prevented from allowing most or even all words in the document to be considered as signaling them. A simple approach to achieving this is to assume `Strong Nuclearity', relying on Marcu's (\citeyear{Marcu1996}) Compositionality Criterion for Discourse Trees (CCDT), which suggests that if a relation holds between two blocks of EDUs, then it also holds between their head EDUs. While this simplification may not be entirely accurate in all cases, Table \ref{tab:taxonomy} suggests that it captures most signals, and allows us to reduce the space of candidate signal tokens to just the two head EDUs implicated in a relation. We will refer to signals within the head units of a relation as `endocentric' and signals outside this region as `exocentric'. Figure \ref{fig:strong_nunclearity} illustrates this, where units [64] and [65] are the respective heads of two blocks of EDUs, and unit [65] in fact contains a plausible endocentric signal for the \textsc{result} relation, the discourse marker \textit{thus}.

More problematic caveats for the frequentist approach are the potential for over/underfitting and ambiguity. The issue of overfitting is especially thorny in small datasets, in which certain content words appear coincidentally in discourse segments with a certain function. Table \ref{tab:freq_based} shows the most distinctive lexical types for several discourse relations in GUM based on pure ratio of occurrence in head EDUs marked for those relations. On the left, types are chosen which have a maximal frequency in the relevant relationship compared with their overall frequency in the corpus. This quickly overfits the contents of the corpus, selecting irrelevant words such as \textit{holiest} and \textit{Slate} for the \textsc{circumstance} relation, or \textit{hypnotizing} and \textit{currency} for \textsc{concession}. The same lack of filtering can, however, yield some potentially relevant lexical items, such as \textit{causing} for \textsc{result}  or even highly specific content words such as \textit{ammonium}, which are certainly not discourse markers, but whose appearance in a \textsc{sequence} is not accidental: the word is in this case typical for sequences in how-to guides, where use of ingredients in a recipe is described in a sequence. Even if these kinds of items may be undesirable candidates for signal words in general, it seems likely that some rare content words may function as signals in context, such as evaluative adjectives (e.g.~\textit{exquisite}) enabling readers to recognize an \textsc{evaluation}.\footnote{Arguably adjectives are the strongest signals of both positive and negative evaluative language (cf.~\citealt{BenamaraTaboadaMathieu2017}).}

\begin{table}[t]
\centering
\begin{tabular}{p{2.6cm}|>{\raggedright\arraybackslash}p{5.5cm}|>{\raggedright\arraybackslash}p{5.5cm}}
\hline

relation & $f>0$  & $f>10$ \\
\hline
\textsc{solutionhood} & viable, contributed, 60th, touched, Palestinians & What, ?, Why, did, How \\
\textsc{circumstance} & holiest, Eventually, fell, Slate, transition & October, When, Saturday, After, Thursday \\
\textsc{result} & minuscule, rebuilding, distortions, struggle, causing & result, Phoenix, wikiHow, funny, death \\
\textsc{concession} & Until, favoured, hypnotizing, currency, curiosity & Although, While, though, However, call \\
\textsc{justify} & payoff, skills, net, Presidential, supporters & NATO, makes, simply, Texas, funny \\
\textsc{sequence} & Feel, charter, ammonium, evolving, rests & bottles, Place, Then, baking, soil \\
\textsc{cause} & malfunctioned, jams, benefiting, mandate, recognising & because, wanted, religious, projects, stuff
\end{tabular}
\caption{Most distinctive lexemes for some relations in GUM, with different frequency thresholds.}
\label{tab:freq_based}
\end{table}

If we are willing to give up on the latter kind of rare items, the overfitting problem can be alleviated somewhat by setting a frequency threshold for each potential signal lexeme, thereby suppressing rare items. The items on the right of the table are limited to types occurring more than 10 times. Since the most distinctive items on the left are all comparatively rare (and therefore exclusive to their relations), they do not overlap with the items on the right.

Looking at the items on the right, several signals make intuitive sense, especially for relations such as \textsc{solutionhood} (used for question-answer pairs) or \textsc{concession}, which show the expected WH words and auxiliary \textit{did}, or discourse markers such as \textit{though}, respectively. At the same time, some high frequency items may be spurious, such as \textit{NATO} for \textsc{justify}, which could perhaps be filtered out based on low dispersion across documents, but also \textit{stuff} for \textsc{cause}, which probably could not be.

Another problem with the lists on the right is that some expected strong signals, such as the word \textit{and} for \textsc{sequence} are absent from the table. This is not because \textit{and} is not frequent in sequences, but rather because it is a ubiquitous word, and as a result, it is not very specific to the relation. However if we look at actual examples of \textit{and} inside and outside of sequences, it is easy to notice that the kind of \textit{and} that does signal a relation in context is often clause initial as in \ref{ex:seqinitial_and} and very different from the adnominal coordinating \textit{and}s in \ref{ex:adnominal_and}, which do not signal the relation:

\ex. [she was made a Dame by Elizabeth II for services to architecture,] [\textbf{and} in 2015 she became the first and only woman to be awarded the Royal Gold Medal]$_{\textsc{sequence}}$ \label{ex:seqinitial_and}

\ex. [Gordon visited England \textbf{and} Scotland in 1686.] [In 1687 \textbf{and} 1689 he took part in expeditions against the Tatars in the Crimea]$_{\textsc{sequence}}$ \label{ex:adnominal_and}

These examples suggest that a data-driven approach to signal detection needs some way of taking context into account. In particular, we would like to be able to compare instances of signals and quantify how strong the signal is in each case. In the next section, we will attempt to apply a neural model with contextualized word embeddings \citep{PetersNeumannIyyerEtAl2018} to this problem, which will be capable of learning contextualized representations of words within the discourse graph.

\subsection{A Contextualized Neural Model} \label{neuralmodel}

\paragraph{Task and Model Architecture} Since we are interested in identifying unrestricted signaling devices, we deliberately avoid a supervised learning approach as used in automatic signal detection trained on resources such as PDTB. While recent work on PDTB connective detection (\citealt{MullerBraudMorey2019}, \citealt{YuZhuLiuEtAl2019})  achieves good results (F-Scores of around 88-89 for English PDTB explicit connectives), the use of such supervised approaches would not tell us about new signaling devices, and especially about unrestricted lexical signals and other coherence devices not annotated in PDTB. Additionally, we would be restricted to the newspaper text types represented in the Wall Street Journal corpus, since no other large English corpus has been annotated for anchored signals.

Instead, we will adopt a distantly supervised approach: we will task a model with supervised discourse relation classification on data that has not been annotated for signals, and infer the positions of signals in the text by analyzing the model's behavior. A key assumption, which we will motivate below, is that signals can have different levels of \textit{signaling strength}, corresponding to their relative importance in identifying a relation. We would like to assume that different signal strength is in fact relevant to human analysts' decision making in relation identification, though in practice we will be focusing on model estimates of strength, the usefulness of which will become apparent below.

\begin{figure*}[h!tb]
\centering
\includegraphics[width=\textwidth, trim={1.5cm 11cm 0.5cm 4.2cm}, clip]{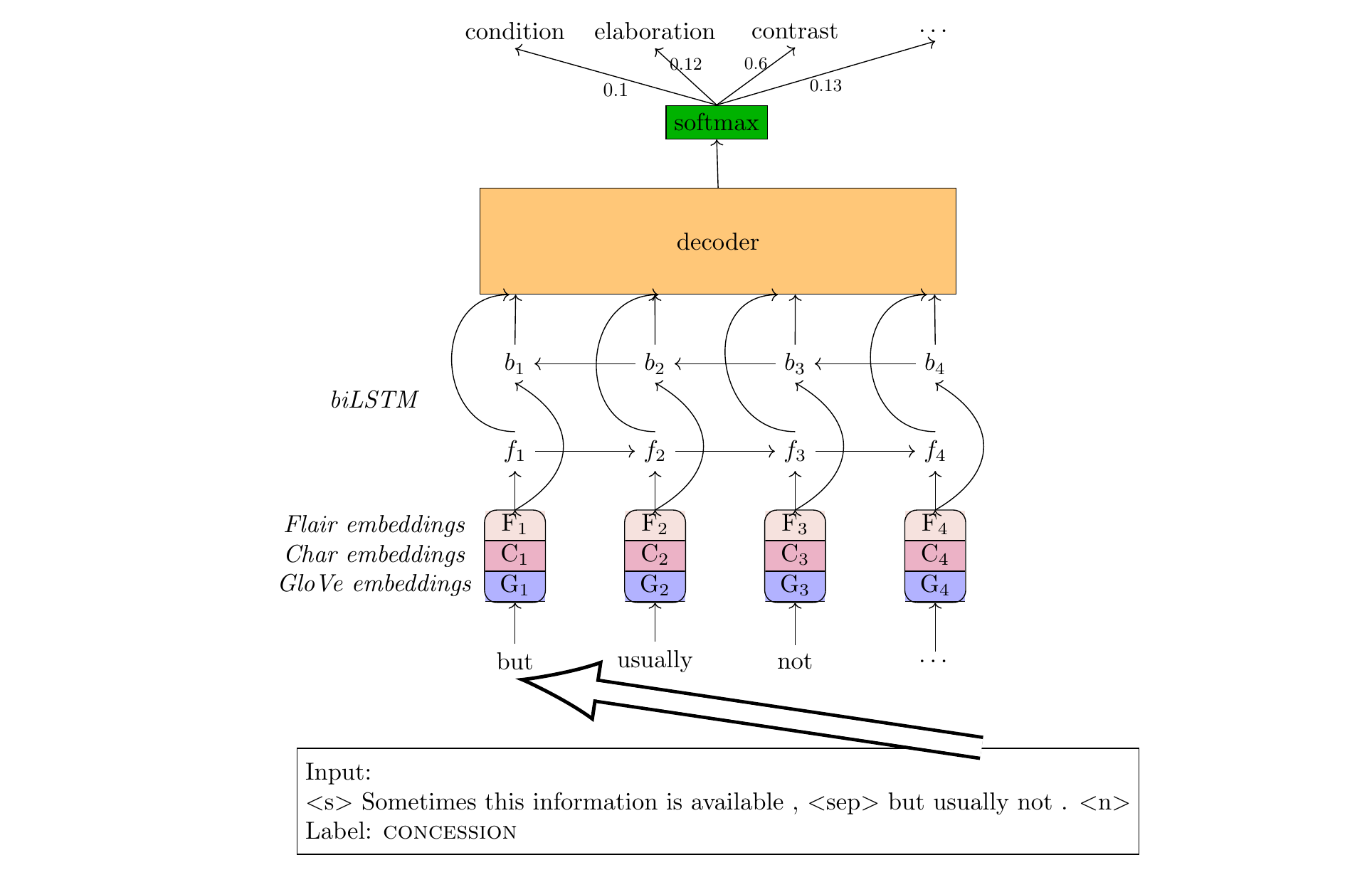}
\caption{Model Architecture. EDU pairs are fed along with satellite, nucleus and separator markers in text order to an encoder using GloVe, Flair and character embeddings. A biLSTM classifier outputs probabilities for the possible discourse relations.}
\label{fig:model_architecture}
\par\vspace{-15pt}\par
\end{figure*}

As a framework, we use the sentence classifier configuration of FLAIR \citep{AkbikBergmannVollgraf2019} with a biLSTM encoder/classifier architecture fed by character and word level representations composed of a concatenation of fixed 300 dimensional GloVe embeddings \citep{PenningtonSocherManning2014}, pre-trained contextualized FLAIR word embeddings, and pre-trained contextualized character embeddings from AllenNLP \citep{GardnerGrusNeumannEtAl2018} with FLAIR's default hyperparameters. The model's architecture is shown in Figure \ref{fig:model_architecture}.

Contextualized embeddings \citep{PetersNeumannIyyerEtAl2018} have the advantage of giving distinct representations to different instances of the same word based on the surrounding words, meaning that an adnominal \textit{and} connecting two NPs can be distinguished from one connecting two verbs based on its vector space representation in the model. Using character embeddings, which give vector space representations to substrings within each word, means that the model can learn the importance of morphological forms, such as the English gerund's \textit{-ing} suffix, even for out-of-vocabulary items not seen during training.

Formally, the input to our system is formed of EDU pairs which are the head units within the respective blocks of discourse units that they belong to, which are in turn connected by an instance of a discourse relation.\footnote{To simplify evaluation and ensure compatibility between the unit of measurement for human and model signal detection, we use gold segmented EDUs as input. An examination of the model's performance in conjunction with automatic EDU segmentation is outside the scope of this study.} This means that every discourse relation in the corpus is expressed as exactly one EDU pair. Each EDU is encoded as a (possibly padded) sequence of $n$-dimensional vector representations of each word ${x_1,..,x_T}$, with some added separators which are encoded in the same way and described below. The bidirectional LSTM composes representations and context for the input, and a fully connected softmax layer gives the probability of each relation:
\begin{align*}
softmax(rel_i) = \frac{e^{rel_i}}{\sum_{j=1}^K e^{rel_j}}, \\
h_t^\delta = f_h(x_{t-1},h_{t-1},c_{t-1};\theta)), \\
c_t^\delta = f_c(x_{t-1},h_{t-1},c_{t-1};\theta))
\end{align*}
where the probability of each relation $rel_i$ is derived from the composed output of the function $h$ across time steps $0 \ldots t$, $\delta \in \{b,f\}$ is the direction of the respective LSTMs, $c_t^\delta$ is the recurrent context in each direction and $\theta = {W,b}$ gives the model weights and bias parameters (see \citealt{AkbikBergmannVollgraf2019} for details). Note that although the output of the system is ostensibly a probability distribution over relation types, we will not be directly interested in the most probable relation as outputted by the classifier, but rather in analyzing the model's behavior with respect to the input word representations as potential signals of each relation.

In order to capitalize on the system's natural language modeling knowledge, EDU satellite-nucleus pairs are presented to the model in text order (i.e.~either the nucleus or the satellite may come first).\footnote{Flair embeddings compute a pooled word representation from contextualized character embeddings and previous occurrences of the entire word string \cite{AkbikBergmannVollgraf2019}, meaning that the pre-trained model retains knowledge and expectations regarding the order of words in a text.} However, the model is given special separator symbols indicating the positions of the satellite and nucleus, which are essential for deciding the relation type (e.g.~\textsc{cause} vs.~\textsc{result}, which may have similar cue words but lead to opposite labels), and a separator symbol indicating the transition between satellite and nucleus. This setup is illustrated in \ref{ex:neural_input}. 

\ex. $<$s$>$ Sometimes this information is available , $<$sep$>$ but usually not . $<$n$>$\\
Label: \textsc{concession}\label{ex:neural_input}

In this example, the satellite precedes the nucleus and is therefore presented first. The model is made aware of the fact that the segment on the left is the satellite thanks to the tag \texttt{<s>}. Since the LSTM is bi-directional, it is aware of positions being within the nucleus or satellite, as well as their proximity to the separator, at every time step.
We reserve the signal-annotated subset of 12 documents from GUM for testing, which contains 1,185 head EDU pairs (each representing one discourse relation), and a random selection of 12 further documents from the remaining RST-annotated GUM data (1,078 pairs) is taken as development data, leaving 102 documents (5,828 pairs) for training. The same EDUs appear in multiple pairs if a unit has multiple children with distinct relations, but no instances of EDUs are shard across partitions, since the splits are based on document boundaries. We note again that for the training and development data, we have no signaling annotations of any kind; this is possible since the network does not actually use the human signaling annotations we will be evaluating against: its distant supervision consists solely of the RST relation labels. 

\paragraph{Relation Classification Performance} Although only used as an auxiliary training task, we can look at the model's performance on predicting discourse relations, which is given in Table \ref{tab:model_performance}. Unsurprisingly, the model performs best on the most frequent relations in the corpus, such as \textsc{elaboration} or \textsc{joint}, but also on rarer ones which tend to be signaled explicitly, such as \textsc{condition} (often signaled explicitly by \textit{if}), \textsc{solutionhood} (used for question-answer pairs signaled by question marks and WH words), or \textsc{concession} (DMs such as \textit{although}). However, the model also performs reasonably well for some trickier (i.e.~less often introduced by unambiguous DMs) but frequent relations, such as \textsc{preparation}, \textsc{circumstance}, and \textsc{sequence}. Rare relations with complex contextual environments, such as \textsc{result}, \textsc{justify} or \textsc{antithesis}, unsurprisingly do not perform well, with the latter two showing an F-score of 0. The relation \textsc{restatement}, which also shows no correct classifications, reveals a weakness of the model: while it is capable of recognizing signals in context, it cannot learn that repetition in and of itself, regardless of specific areas in vector space, is important (see Section \ref{discussion} for more discussion of these and other classification weaknesses).

\begin{table}[h!tb]
\centering
\begin{tabular}{l|r|r|r|r|r|r|r}
\hline
\textbf{relation} & \textbf{N train} & \textbf{\% train} & \textbf{N test} & \textbf{\% test} & \textbf{P} & \textbf{R} & \textbf{F} \\
\hline
\textsc{antithesis} & 87 & 1.49 & 22 & 2.88 & 0 & 0 & 0 \\
\textsc{background} & 406 & 6.97 & 58 & 7.59 & 46.67 & 36.21 & 40.78 \\
\textsc{cause} & 224 & 3.84 & 18 & 2.36 & 23.08 & 16.67 & 19.36 \\
\textsc{circumstance} & 291 & 4.99 & 23 & 3.01 & 34.69 & 73.91 & 47.22 \\
\textsc{concession} & 237 & 4.07 & 34 & 4.45 & 20.00 & 23.53 & 21.62 \\
\textsc{condition} & 180 & 3.09 & 30 & 3.93 & 85.71 & 80.00 & 82.76 \\
\textsc{contrast} & 216 & 3.71 & 34 & 4.45 & 14.00 & 20.59 & 16.67 \\
\textsc{elaboration} & 921 & 15.8 & 116 & 15.18 & 34.85 & 39.66 & 37.10 \\
\textsc{evaluation} & 252 & 4.32 & 31 & 4.06 & 36.36 & 12.90 & 19.04 \\
\textsc{evidence} & 131 & 2.25 & 16 & 2.09 & 23.08 & 18.75 & 20.69 \\
\textsc{joint} & 1083 & 18.58 & 165 & 21.6 & 56.48 & 73.94 & 64.04 \\
\textsc{justify} & 168 & 2.88 & 35 & 4.58 & 0 & 0 & 0 \\
\textsc{motivation} & 112 & 1.92 & 12 & 1.57 & 13.33 & 16.67 & 14.81 \\
\textsc{preparation} & 432 & 7.41 & 68 & 8.9 & 69.33 & 76.47 & 72.73 \\
\textsc{purpose} & 117 & 2.01 & 19 & 2.49 & 83.33 & 78.95 & 81.08 \\
\textsc{restatement} & 119 & 2.04 & 31 & 4.06 & 0 & 0 & 0 \\
\textsc{result} & 167 & 2.87 & 18 & 2.36 & 10 & 5.56 & 7.15 \\
\textsc{sequence} & 530 & 9.09 & 16 & 2.09 & 11.76 & 12.50 & 12.12 \\
\textsc{solutionhood} & 155 & 2.66 & 18 & 2.36 & 54.55 & 66.67 & 60.00 \\
\hline
\textbf{total (micro avg)} & \textbf{5828} & \textbf{100} & \textbf{764} & \textbf{100} & \textbf{44.37} & \textbf{44.37} & \textbf{44.37} \\
\hline
  \quad\quad~(macro avg) &  &  &  &  & 32.49 & 34.37 & 32.48 \\
\end{tabular}
\caption{Model performance on relation classification in the test set.}
\label{tab:model_performance}
\end{table}

Although this is not the actual task targeted by the current paper, we may note that the overall performance of the model, with an F-Score of 44.37, is not bad, though below the performance of state-of-the-art full discourse parsers (see \citealt{MoreyMullerAsher2017}) -- this is to be expected, since the model is not aware of the entire RST tree, rather looking only at EDU pairs out of context, and given that standard scores on RST-DT come from a larger and more homogeneous corpus, with with fewer relations and some easy cases that are absent from GUM.\footnote{For example, RST-DT annotates relative clauses, which are often easy to identify, as \textsc{elaboration}s, while version 5 of GUM, used here, does not segment adnominal clauses as EDUs at all. In the newest version 6 of GUM, released during the review period, the same segmentation guidelines are followed as in RST-DT. Additionally note that most discourse parsing on RST-DT uses a collapsed set of only 16 relations, as opposed to the 20 used here. However, we would like to emphasize that discourse parsers have entirely different goals, so that these numbers and parsing scores would be apples and oranges even if the same underlying corpus were used.} 

Given the model's performance on relation classification, which is far from perfect, one might wonder whether signal predictions made by our analysis should be trusted. This question can be answered in two ways: first, quantitatively, we will see in Section \ref{evaluation} that model signal predictions overlap considerably with human judgments, even when the predicted relation is incorrect. Intuitively, for similar relations, such as \textsc{concession} or \textsc{contrast}, both of which are adversative, the model may notice a relevant cue (e.g.~`but', or contrasting lexical items) despite choosing the wrong one. Second, as we will see below, we will be analyzing the model's behavior with respect to the probability of the correct relation, regardless of the label it ultimately chooses, meaning that the importance of predicting the correct label exactly will be diminished further.

\paragraph{Signaling Metric} The actual performance we are interested in evaluating is the model's ability to extract signals for given discourse relations, rather than its accuracy in predicting the relations. To do so, we must extract anchored signal predictions from the model, which is non-trivial. While earlier work on interpreting neural models has focused on token-wise softmax probability \citep{Zeldes2018} or attention weights \citep{GhaeiniFernTadepalli2018}, using contextualized embeddings complicates the evaluation: since word representations are adjusted to reflect neighboring words, the model may assign higher importance to the word standing next to what a human annotator may interpret as a signal. Example \ref{ex:to_provide1} illustrates the problem:

\ex. \textcolor[RGB]{230, 230, 230}{To} \textbf{\textcolor[RGB]{53, 53, 53}{provide}} \textcolor[RGB]{165, 165, 165}{information} \textcolor[RGB]{179, 179, 179}{on} \textcolor[RGB]{175, 175, 175}{the} \textcolor[RGB]{160, 160, 160}{analytical} \textcolor[RGB]{157, 157, 157}{sample} \textcolor[RGB]{187, 187, 187}{as} \textcolor[RGB]{170, 170, 170}{a} \textcolor[RGB]{168, 168, 168}{whole} \textcolor[RGB]{207, 207, 207}{,}  $\xrightarrow[\text{pred:preparation}]{\text{gold:purpose}}$ \textcolor[RGB]{168, 168, 168}{two} \textcolor[RGB]{170, 170, 170}{additional} \textcolor[RGB]{164, 164, 164}{demographic} \textcolor[RGB]{175, 175, 175}{variables} \textcolor[RGB]{182, 182, 182}{are} \textcolor[RGB]{165, 165, 165}{included} \textcolor[RGB]{230, 230, 230}{.} 
 \label{ex:to_provide1}

Each word in \ref{ex:to_provide1} is shaded based on the softmax probability assigned to the correct relation of the satellite, i.e.~how `convincing' the model found the word in terms of local probability. In addition, the top-scoring word in each sentence is rendered in boldface for emphasis. The gold label for the relation is placed above the arrow, which indicates the direction of the relation (satellite to nucleus), and the model's predicted label is shown under the arrow.

Intuitively, the strongest signal of the \textsc{purpose} relation in \ref{ex:to_provide1} is the initial infinitive marker \textit{To} -- however, the model ranks the adjacent \textit{provide} higher and almost ignores \textit{To}. We suspect that the reason for this, and many similar examples in the model evaluated based on relation probabilities, is that contextual embeddings allow for a special representation of the word \textit{provide} next to \textit{To}, making it difficult to tease apart the locus of the most meaningful signal.

To overcome this complication, we use the logic of permutation importance, treating the neural model as a black box and manipulating the input to discover relevant features in the data (cf.~\citealt{CasalicchioMolnarBischl2019}). We reason that this type of evaluation is more robust than, for example, examining model internal attention weights because such weights are not designed or trained with a reward ensuring they are informative -- they are simply trained on the same classification error loss as the rest of the model.\footnote{We thank an anonymous reviewer for pointing out \cite{WiegreffePinter2019}, who offer experimental support for the variable utility of attention as an explanatory metric.} Instead, we can withhold potentially relevant information from the model directly: After training is complete, we feed the test data to the model in two forms -- as-is, and with each word masked, as shown in \ref{ex:mask}.

\ex. \begin{tabbing} Original:\quad  \= $<$s$>$ \= \ To\quad\  \= provide \= information \= ... \= $<$sep$>$ \= ... \= $<$n$>$ \kill 
 Original: \> $<$s$>$ \> \ To \> provide \> information \> ... \> $<$sep$>$ \> ... \> $<$n$>$ \\       
 Masked1: \> $<$s$>$ \> $<$X$>$ \> provide \> information \> ... \> $<$sep$>$ \> ... \> $<$n$>$ \\ 
     Masked2: \> $<$s$>$ \> \ To \> \ $<$X$>$ \> information \> ... \> $<$sep$>$ \> ... \> $<$n$>$ \\
   Masked3: \> $<$s$>$ \> \ To \> provide \> \ $<$X$>$ \> ... \> $<$sep$>$ \> ... \> $<$n$>$ 
   \end{tabbing}
  \glt  Label: \textsc{purpose} \label{ex:mask}

We reason that, if a token is important for predicting the correct label, masking it will degrade the model's classification accuracy, or at least reduce its reported classification certainty.\footnote{An anonymous reviewer has suggested that part of the model's response to masking may be the result not only of missing discourse signals, but also of ungrammaticality due to missing words. While this could be the case to some extent, we would like to point out that the model should be somewhat accustomed to the masking situation since masked tokens are merely represented as OOV word embeddings (which occur regularly at train and test time), and because drop out is applied to the model during training, meaning that the model is constantly exposed to some missing information. If the masked information were not crucial to relation classification, the model should still work correctly, whereas if an ungrammatical construction is truly preventing the identification of the relation, then that construction may well have been a signal, and the masking approach would be doing its job as intended.} In \ref{ex:mask}, it seems reasonable to assume that masking the word `To' has a greater impact on predicting the label \textsc{purpose} than masking the word `provide', and even less so, the following noun `information'. We therefore use \textit{reduction in softmax probability of the correct relation} as our signaling strength metric for the model. We call this metric ${\Delta}_s$ (for delta-softmax), which can be written as:
\begin{align*}
    \Delta _s^{rel,t_i} = softmax(rel|X_{mask=\phi}) - softmax(rel|X_{mask=i})
\end{align*}
where $rel$ is the true relation of the EDU pair, $t_i$ represents the token at index $i$ of $N$ tokens, and $X_{mask=i}$ represents the input sequence with the masked position $i$ (for $i \in 1 \ldots N$ ignoring separators, or $\phi$, the empty set).

To visualize the model's predictions, we compare ${\Delta}_s$ for a particular token to two numbers: the maximum ${\Delta}_s$ achieved by any token in the current pair (a measure of relative importance for the current classification) and the maximum ${\Delta}_s$ achieved by any token in the current document (a measure of how strongly the current relation is signaled compared to other relations in the text). We then shade each token 50\% based on the first number and 50\% based on the second. As a result, the most valid cues in an EDU pair are darker than their neighbors, but EDU pairs with no good cues are overall very light, whereas pairs with many good signals are darker. Some examples of this visualization are given in \ref{ex:to_provide2}-\ref{ex:unlikely} (human annotated endocentric signal tokens are marked by double underlines).

\ex. \doubleunderline{\textbf{\textcolor[RGB]{61, 61, 61}{To}}} \textcolor[RGB]{112, 112, 112}{provide} \textcolor[RGB]{205, 205, 205}{information} \textcolor[RGB]{230, 230, 230}{on} \textcolor[RGB]{230, 230, 230}{the} \textcolor[RGB]{230, 230, 230}{analytical} \textcolor[RGB]{230, 230, 230}{sample} \textcolor[RGB]{230, 230, 230}{as} \textcolor[RGB]{230, 230, 230}{a} \textcolor[RGB]{230, 230, 230}{whole} \textcolor[RGB]{230, 230, 230}{,}  $\xrightarrow[\text{pred:preparation}]{\text{gold:purpose}}$ \textcolor[RGB]{230, 230, 230}{two} \textcolor[RGB]{183, 183, 183}{additional} \textcolor[RGB]{230, 230, 230}{demographic} \textcolor[RGB]{230, 230, 230}{variables} \textcolor[RGB]{94, 94, 94}{are} \textcolor[RGB]{194, 194, 194}{included} \textcolor[RGB]{163, 163, 163}{.}  \label{ex:to_provide2}

\ex. \textcolor[RGB]{230, 230, 230}{Telling} \textcolor[RGB]{230, 230, 230}{good} \textcolor[RGB]{230, 230, 230}{jokes} \textcolor[RGB]{230, 230, 230}{is} \textcolor[RGB]{230, 230, 230}{an} \textcolor[RGB]{230, 230, 230}{art} \textcolor[RGB]{230, 230, 230}{that} \doubleunderline{\textcolor[RGB]{230, 230, 230}{comes} \textcolor[RGB]{230, 230, 230}{naturally}} \textcolor[RGB]{230, 230, 230}{to} \doubleunderline{\textcolor[RGB]{230, 230, 230}{some} \textcolor[RGB]{211, 211, 211}{people}} \textcolor[RGB]{135, 135, 135}{,}  $\xleftarrow[\text{pred:contrast}]{\text{gold:contrast}}$ \doubleunderline{\textbf{\textcolor[RGB]{21, 21, 21}{but}}} \textcolor[RGB]{209, 209, 209}{for} \doubleunderline{\textcolor[RGB]{207, 207, 207}{others}} \textcolor[RGB]{230, 230, 230}{it} \doubleunderline{\textcolor[RGB]{217, 217, 217}{takes} \textcolor[RGB]{230, 230, 230}{practice} \textcolor[RGB]{230, 230, 230}{and} \textcolor[RGB]{189, 189, 189}{hard} \textcolor[RGB]{230, 230, 230}{work}} \textcolor[RGB]{230, 230, 230}{.} \label{ex:telling}

\ex. \textcolor[RGB]{230, 230, 230}{It} \textcolor[RGB]{230, 230, 230}{is} \textcolor[RGB]{230, 230, 230}{possible} \textcolor[RGB]{230, 230, 230}{that} \textcolor[RGB]{230, 230, 230}{these} \textcolor[RGB]{230, 230, 230}{two} \textcolor[RGB]{230, 230, 230}{children} \textcolor[RGB]{230, 230, 230}{understood} \textcolor[RGB]{230, 230, 230}{the} \textcolor[RGB]{230, 230, 230}{task} \textcolor[RGB]{230, 230, 230}{and} \textcolor[RGB]{230, 230, 230}{really} \textcolor[RGB]{230, 230, 230}{did} \textcolor[RGB]{230, 230, 230}{believe} \textcolor[RGB]{230, 230, 230}{that} \textcolor[RGB]{230, 230, 230}{the} \textcolor[RGB]{230, 230, 230}{puppet} \textcolor[RGB]{230, 230, 230}{did} \textcolor[RGB]{230, 230, 230}{not} \textcolor[RGB]{230, 230, 230}{produce} \textcolor[RGB]{230, 230, 230}{any} \textcolor[RGB]{230, 230, 230}{poor} \textcolor[RGB]{230, 230, 230}{descriptions} \textcolor[RGB]{230, 230, 230}{,} \textcolor[RGB]{230, 230, 230}{and} \textcolor[RGB]{230, 230, 230}{in} \textcolor[RGB]{230, 230, 230}{this} \textcolor[RGB]{230, 230, 230}{regard} \textcolor[RGB]{230, 230, 230}{,} \textcolor[RGB]{230, 230, 230}{are} \textcolor[RGB]{230, 230, 230}{not} \textcolor[RGB]{230, 230, 230}{yet} \textcolor[RGB]{230, 230, 230}{adult-like} \textcolor[RGB]{230, 230, 230}{in} \textcolor[RGB]{230, 230, 230}{their} \textcolor[RGB]{230, 230, 230}{SI} \textcolor[RGB]{230, 230, 230}{interpretations} \textcolor[RGB]{230, 230, 230}{.}  $\xleftarrow[\text{pred:evaluation}]{\text{gold:evaluation}}$ \doubleunderline{\textcolor[RGB]{230, 230, 230}{This}} \textcolor[RGB]{230, 230, 230}{is} \doubleunderline{\textbf{\textcolor[RGB]{41, 41, 41}{unlikely}}} \label{ex:unlikely}

The highlighting in \ref{ex:to_provide2} illustrates the benefits of the masking based evaluation compared to \ref{ex:to_provide1}: the token \textit{To} is now clearly the strongest signal, and the verb is taken to be less important, followed by the even less important object of the verb. This is because removing the initial \textit{To} hinders classification much more than the removal of the verb or noun. We note also that although the model in fact misclassified this example as \textsc{preparation}, we can still use masking importance to identify \textit{To}, since the score queried from the model corresponds to a relative decrease in the probability of the \textit{correct} relation, \textsc{purpose}, even if this was not the highest scoring relation overall.

In \ref{ex:telling} we see the model's ability to correctly predict \textsc{contrast} based on the DM \textit{but}. Note that despite a rather long sentence, the model does not need any other word nearly as much for the classification. Although the model is not trained explicitly to detect discourse markers, the DM can be recognized due to the fact that masking it leads to a drop of 66\% softmax probability (${\Delta}_s$=0.66) of this pair representing the \textsc{contrast} relation. We can also note that a somewhat lower scoring content word is also marked: \textit{hard} (${\Delta}_s$=0.18). In our gold signaling annotations, this word was marked together with \textit{comes naturally} as a signal, due to the contrast between the two concepts (additionally, \textit{some people} is flagged as a signal along with \textit{others}). The fact that the model finds \textit{hard} helpful, but does not need the contextual near antonym \textit{naturally}, suggests that it is merely learning that words in the semantic space near \textit{hard} may indicate \textsc{contrast}, and not learning about the antonymous relationship -- otherwise we would expect to see `naturally' have a stronger score (see also the discussion in Section \ref{discussion}).

Finally \ref{ex:unlikely} shows that, much like in the case of \textit{hard}, the model is not biased towards traditional DMs, confirming that it is capable of learning about content words, or neighborhoods of content words in vector space. In a long EDU pair of 41 words, the model relies almost exclusively on the word \textit{unlikely} (${\Delta}_s$=0.36) to correctly label the relation as \textsc{evaluation}. By contrast, the anaphoric demonstrative `This' flagged by the human annotator, which is a more common function word, is disregarded, perhaps because it can appear with several other relations, and is not particularly exclusive to \textsc{evaluation}. These results suggest that the model may be capable of recognizing signals through distant supervision, allowing it to validate human annotations, to potentially point out signals that may be missed by annotators, and most importantly, to quantify signaling strength on a sliding scale. At the same time, we need a way to evaluate the model's quality and assess the kinds of errors it makes, as well as what we can learn from them. We therefore move on to evaluating the model and its errors next.

\section{Evaluation and Error Analysis} \label{evaluation}

\paragraph{Evaluation Metric} To evaluate the neural model, we would like to know how well ${\Delta}_s$ corresponds to annotators' gold standard labels. This leads to two kinds of problems: the first is that the model is distantly supervised, and therefore does not know about signal types, subtypes, or any aspect of signaling annotation and its relational structure. The second problem is that signaling annotations are categorical, and do not correspond to the ratio-scaled predictions provided by ${\Delta}_s$ (this is in fact one of the motivations for desiring a model-based estimate of signaling strength). 

The first issue means that we can only examine the model's ability to locate signals -- not to classify them. Although there may be some conceivable ways of analyzing model output to identify classes such as DMs (which are highly lexicalized, rather than representing broad regions of vector space, as words such as \textit{unlikely} might), or more contextual relational signals, such as pronouns, this line of investigation is beyond the scope of the present paper. A naive solution to the second problem might be to identify a cutoff point, e.g.~deciding that all and only words scoring ${\Delta}_s>$0.15 are predicted to be signals.

The problem with the latter approach is that sentences can be very different in many ways, and specifically in both length and in levels of ambiguity. Sentences with multiple, mutually redundant cues, may produce lower ${\Delta}_s$ scores compared to shorter sentences with a subset of the same cues. Conversely, in very short sentences with low signal strength, the model may reasonably be expected to degrade very badly with the deletion of almost any word, as the context becomes increasingly incomprehensible.

\begin{figure*}[h!tb]
\centering
\includegraphics[width=\textwidth]{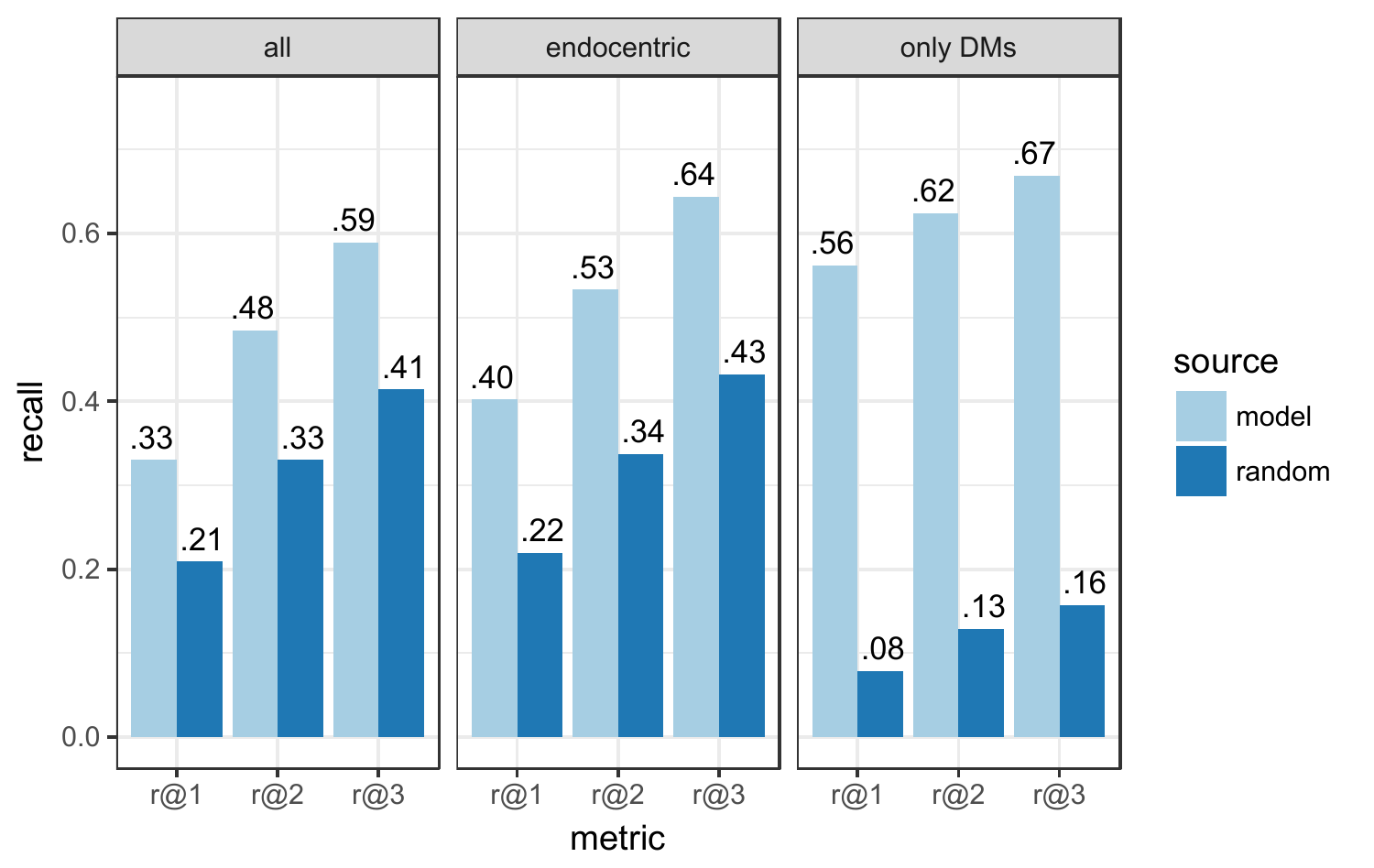}
\caption{Signal recall with 1, 2 or 3 guesses for the neural model and a random guess baseline. Left: all signals included; middle: only endocentric cases, restricted to head EDUs shown to the model; right: only discourse markers (DMs).}
\label{fig:recallk}
\par\vspace{-15pt}\par
\end{figure*}

For these reasons, we choose to adopt an evaluation metric from the paradigm of information retrieval, and focus on \textit{recall@k} (recall at rank \textit{k}, for $k=1,2,3$...). The idea is to poll the model for each sentence in which some signals have been identified, and see whether the model is able to find them if we let it guess using the word with the maximal ${\Delta}_s$ score (recall@1), regardless of how high that score is, or alternatively relax the evaluation criteria and see whether the human annotator's signal tokens appear at rank 2 or 3. Figure \ref{fig:recallk} shows numbers for recall@k for the top 3 ranks outputted by the model, next to random guess baselines. 

\begin{figure*}[hbt]
\centering
\includegraphics[width=75mm]{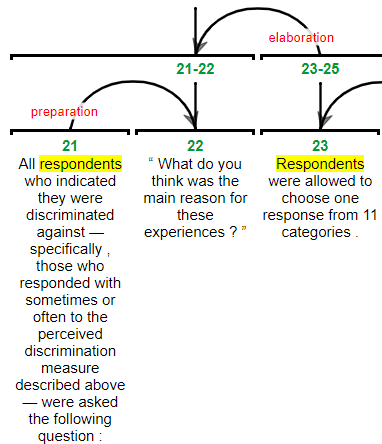}
\caption{Exocentric signal not detectable by the model. The \textsc{elaboration} pointing from unit [23] to [22] is signaled by a coreferential phrase appearing in another satellite of [22].}
\label{fig:exocentric}
\par\vspace{-10pt}\par
\end{figure*}

The left, middle and right panels in Figure \ref{fig:recallk} correspond to measurements when all signals are included, only cases contained entirely in the head EDUs shown to the model, and only DMs, respectively. The scenario on the left is rather unreasonable and is included only for completeness: here the model is also penalized for not detecting signals such as lexical chains, part of which is outside the units that the model is being shown. An example of such a case can be seen in Figure \ref{fig:exocentric}. The phrase \textit{Respondents} in unit [23] signals the relation \textsc{elaboration}, since it is coreferential with a previous mention of the respondents in [21]. However, because the model is only given heads of EDU blocks to classify, it does not have access to the first occurrence of \textit{respondents} while predicting the \textsc{elaboration} relation -- the first half of the signal token set is situated in a child of the nucleus EDU before the relation, i.e.~it belongs to group IV in the taxonomy in Table \ref{tab:taxonomy}. Realistically, our model can only be expected to learn about signals from `directly participating' EDUs, i.e.~groups I, II, VI and VII, the `endocentric' signal groups from Section \ref{signal_taxonomy}. 

Although most signals belong to endocentric categories (71.62\% of signaled relations belong to these groups, cf.~Table \ref{tab:taxonomy}), exocentric cases form a substantial portion of signals which we have little hope of capturing with the architecture used here. As a result, recall metrics in the `all signals' scenario are closest to the random baselines, though the signals detected in other instances still place the model well above the baseline. 

A more reasonable evaluation is the one in the middle panel of Figure \ref{fig:recallk}, which includes only endocentric signals as defined in the taxonomy. EDUs with no endocentric signals are completely disregarded in this scenario, which substantially reduces the number of tokens considered to be signals, since, while many tokens are part of some meaningful lexical chain in the document, requiring signals to be contained only in the pair of head units eliminates a wide range of candidates. Although the random baseline is actually very slightly higher (perhaps because eliminated EDUs were often longer ones, sharing small amounts of material with larger parts of the text, and therefore prone to penalizing the baseline; many words mean more chances for a random guess to be wrong), model accuracy is substantially better in this scenario, reaching a 40\% chance of hitting a signal with only one guess, exceeding 53\% with two guesses, and capping at 64\% for recall@3, over 20 points above baseline.\footnote{It should also be noted that a maximum score of 100\% is not a reasonable expectation, since even humans disagree on signal tokens. As a ceiling figure representing human performance we can take the average mutual human recall score for signal tokens in Gessler et al.'s (\citeyear{gessler-etal-2019-discourse}) data, a score of .84.}

Finally, the right panel in the figure shows recall when only DMs are considered. In this scenario, a random guess fares very poorly, since most words are not DMs. The model, by contrast, achieves the highest results in all metrics, since DMs have the highest cue validity for relation classification, and the model attends to them most strongly. With just one guess, recall is over 56\%, and goes as high as 67\% for recall@3. The baseline only goes as high as 16\% for three guesses.

\paragraph{Qualitative Analysis} Looking at the model's performance qualitatively, it is clear that it can detect not only DMs, but also morphological cues (e.g.~gerunds as markers of elaboration, as in \ref{ex:gerund}), semantic classes and sentiment, such as positive and negative evaluatory terms in \ref{ex:sentiment}, as well as multiple signals within the same EDU, as in \ref{ex:multiple}. In fact, only about 8.3\% of the tokens correctly identified by the model in Table \ref{tab:sig_errtypes} below are of the DM type, whereas about 7.2\% of all tokens flagged by human annotators were DMs, meaning that the model frequently matches non-DM items to discourse relation signals (see Performance on Signal Types below). It should also be noted that signals can be recognized even when the model misclassifies relations, since ${\Delta}_s$ does not rely on correct classification: it merely quantifies the contribution of a word in context toward the correct label's score. If we examine the influence of each word on the score of the correct relation, that impact should and does still correlate with human judgments based on what the system may tag as the second or third best class to choose.

\ex. \textcolor[RGB]{230, 230, 230}{For} \textcolor[RGB]{230, 230, 230}{the} \textcolor[RGB]{230, 230, 230}{present} \textcolor[RGB]{230, 230, 230}{analysis} \textcolor[RGB]{230, 230, 230}{,} \textcolor[RGB]{230, 230, 230}{these} \textcolor[RGB]{230, 230, 230}{responses} \textcolor[RGB]{230, 230, 230}{were} \textcolor[RGB]{230, 230, 230}{recoded} \textcolor[RGB]{230, 230, 230}{into} \textcolor[RGB]{230, 230, 230}{nine} \textcolor[RGB]{230, 230, 230}{mutually} \textcolor[RGB]{230, 230, 230}{exclusive} \textcolor[RGB]{230, 230, 230}{categories}  $\xleftarrow[\text{pred:elaboration}]{\text{gold:result}}$ \textbf{\textcolor[RGB]{63, 63, 63}{capturing}} \textcolor[RGB]{219, 219, 219}{the} \textcolor[RGB]{230, 230, 230}{following} \textcolor[RGB]{230, 230, 230}{options} \textcolor[RGB]{135, 135, 135}{:} \label{ex:gerund}

\ex. \textcolor[RGB]{185, 185, 185}{Professor} \textcolor[RGB]{219, 219, 219}{Eastman} \textcolor[RGB]{223, 223, 223}{said} \textcolor[RGB]{207, 207, 207}{he} \textcolor[RGB]{194, 194, 194}{is} \textbf{\textcolor[RGB]{64, 64, 64}{alarmed}} \textcolor[RGB]{230, 230, 230}{by} \textcolor[RGB]{230, 230, 230}{what} \textcolor[RGB]{230, 230, 230}{they} \textcolor[RGB]{230, 230, 230}{found} \textcolor[RGB]{230, 230, 230}{.}  $\xrightarrow[\text{pred:preparation}]{\text{gold:evaluation}}$ \textcolor[RGB]{230, 230, 230}{"} \textcolor[RGB]{230, 230, 230}{Pregnant} \textcolor[RGB]{229, 229, 229}{women} \textcolor[RGB]{187, 187, 187}{in} \textcolor[RGB]{230, 230, 230}{Australia} \textcolor[RGB]{98, 98, 98}{are} \textcolor[RGB]{213, 213, 213}{getting} \textcolor[RGB]{230, 230, 230}{about} \textcolor[RGB]{230, 230, 230}{half} \textcolor[RGB]{171, 171, 171}{as} \textcolor[RGB]{159, 159, 159}{much} \textcolor[RGB]{230, 230, 230}{as} \textcolor[RGB]{230, 230, 230}{what} \textcolor[RGB]{155, 155, 155}{they} \textcolor[RGB]{155, 155, 155}{require} \textcolor[RGB]{223, 223, 223}{on} \textcolor[RGB]{214, 214, 214}{a} \textcolor[RGB]{109, 109, 109}{daily} \textcolor[RGB]{176, 176, 176}{basis} \textcolor[RGB]{111, 111, 111}{.} \label{ex:sentiment}

\ex. \textcolor[RGB]{195, 195, 195}{Even} \textcolor[RGB]{230, 230, 230}{so} \textcolor[RGB]{230, 230, 230}{,} \textcolor[RGB]{230, 230, 230}{estimates} \textcolor[RGB]{230, 230, 230}{of} \textcolor[RGB]{230, 230, 230}{the} \textcolor[RGB]{230, 230, 230}{prevalence} \textcolor[RGB]{230, 230, 230}{of} \textcolor[RGB]{230, 230, 230}{perceived} \textcolor[RGB]{230, 230, 230}{discrimination} \textcolor[RGB]{219, 219, 219}{remains} \textcolor[RGB]{230, 230, 230}{rare}  $\xleftarrow[\text{pred:evidence}]{\text{gold:concession}}$ \textcolor[RGB]{111, 111, 111}{At} \textbf{\textcolor[RGB]{63, 63, 63}{least}} \textcolor[RGB]{230, 230, 230}{one} \textcolor[RGB]{230, 230, 230}{prior} \textcolor[RGB]{230, 230, 230}{study} \textcolor[RGB]{230, 230, 230}{by} \textcolor[RGB]{230, 230, 230}{Kessler} \textcolor[RGB]{225, 225, 225}{and} \textcolor[RGB]{230, 230, 230}{colleagues} \textcolor[RGB]{230, 230, 230}{[} \textcolor[RGB]{230, 230, 230}{15} \textcolor[RGB]{161, 161, 161}{]} \textcolor[RGB]{200, 200, 200}{,} \textcolor[RGB]{136, 136, 136}{however} \textcolor[RGB]{222, 222, 222}{,} \textcolor[RGB]{228, 228, 228}{using} \textcolor[RGB]{230, 230, 230}{measures} \textcolor[RGB]{230, 230, 230}{of} \textcolor[RGB]{230, 230, 230}{perceived} \textcolor[RGB]{224, 224, 224}{discrimination} \textcolor[RGB]{217, 217, 217}{in} \textcolor[RGB]{230, 230, 230}{a} \textcolor[RGB]{230, 230, 230}{large} \textcolor[RGB]{218, 218, 218}{American} \textcolor[RGB]{230, 230, 230}{sample} \textcolor[RGB]{230, 230, 230}{,} \textcolor[RGB]{230, 230, 230}{reported} \textcolor[RGB]{230, 230, 230}{that} \textcolor[RGB]{230, 230, 230}{approximately} \textcolor[RGB]{230, 230, 230}{33} \textcolor[RGB]{212, 212, 212}{\%} \textcolor[RGB]{230, 230, 230}{of} \textcolor[RGB]{230, 230, 230}{respondents} \textcolor[RGB]{156, 156, 156}{reported} \textcolor[RGB]{169, 169, 169}{some} \textcolor[RGB]{122, 122, 122}{form} \textcolor[RGB]{168, 168, 168}{of} \textcolor[RGB]{230, 230, 230}{discrimination} \label{ex:multiple}

Unsurprisingly, the model sometimes make sporadic errors in signal detection for which good explanations are hard to find, especially when its predicted relation is incorrect, as in \ref{ex:remarkable}. Here the evaluative adjective \textit{remarkable} is missed in favor of neighboring words such as \textit{agreed} and a subject pronoun, which are not indicative of the \textsc{evaluation} relation in this context but are part of several cohorts of high scoring words. However, the most interesting and interpretable errors arise when ${\Delta}_s$ scores are high compared to an entire document, and not just among words in one EDU pair, in which most or even all words may be relatively weak signals. As an example of such a false positive with high confidence, we can consider \ref{ex:so_your}. In this example, the model correctly assigns the highest score to the DM \textit{so} marking a \textsc{purpose} relation. However, it also picks up on a recurring tendency in how-to guides in which the second person pronoun referring to the reader is often the benefactee of some action, which contributes to the purpose reading and helps to disambiguate \textit{so}, despite not being considered a signal by annotators.

\ex. \textcolor[RGB]{216, 216, 216}{The} \textcolor[RGB]{99, 99, 99}{agreement} \textcolor[RGB]{89, 89, 89}{was} \textcolor[RGB]{230, 230, 230}{that} \textcolor[RGB]{131, 131, 131}{Gorbachev} \textcolor[RGB]{102, 102, 102}{agreed} \textcolor[RGB]{230, 230, 230}{to} \textcolor[RGB]{230, 230, 230}{a} \textcolor[RGB]{230, 230, 230}{quite} \textcolor[RGB]{230, 230, 230}{remarkable} \textcolor[RGB]{125, 125, 125}{concession} \textcolor[RGB]{230, 230, 230}{:}  $\xrightarrow[\text{pred:preparation}]{\text{gold:evaluation}}$ \textbf{\textcolor[RGB]{64, 64, 64}{he}} \textcolor[RGB]{81, 81, 81}{agreed} \textcolor[RGB]{230, 230, 230}{to} \textcolor[RGB]{230, 230, 230}{let} \textcolor[RGB]{220, 220, 220}{a} \textcolor[RGB]{143, 143, 143}{united} \textcolor[RGB]{149, 149, 149}{Germany} \textcolor[RGB]{230, 230, 230}{join} \textcolor[RGB]{83, 83, 83}{the} \textcolor[RGB]{230, 230, 230}{NATO} \textcolor[RGB]{230, 230, 230}{military} \textcolor[RGB]{230, 230, 230}{alliance} \textcolor[RGB]{230, 230, 230}{.} \label{ex:remarkable}

\ex. \textcolor[RGB]{230, 230, 230}{The} \textcolor[RGB]{220, 220, 220}{opening} \textcolor[RGB]{230, 230, 230}{of} \textcolor[RGB]{230, 230, 230}{the} \textcolor[RGB]{230, 230, 230}{joke} \textcolor[RGB]{230, 230, 230}{—} \textcolor[RGB]{230, 230, 230}{or} \textcolor[RGB]{230, 230, 230}{setup} \textcolor[RGB]{230, 230, 230}{—} \textcolor[RGB]{230, 230, 230}{should} \textcolor[RGB]{230, 230, 230}{have} \textcolor[RGB]{230, 230, 230}{a} \textcolor[RGB]{230, 230, 230}{basis} \textcolor[RGB]{230, 230, 230}{in} \textcolor[RGB]{230, 230, 230}{the} \textcolor[RGB]{230, 230, 230}{real} \textcolor[RGB]{200, 200, 200}{world}  $\xleftarrow[\text{pred:purpose}]{\text{gold:purpose}}$ \textbf{\textcolor[RGB]{7, 7, 7}{so}} \textcolor[RGB]{73, 73, 73}{your} \textcolor[RGB]{230, 230, 230}{audience} \textcolor[RGB]{230, 230, 230}{can} \textcolor[RGB]{230, 230, 230}{relate} \textcolor[RGB]{230, 230, 230}{to} \textcolor[RGB]{230, 230, 230}{it} \textcolor[RGB]{230, 230, 230}{,} \label{ex:so_your}

In other cases, the model points out plausible signals which were passed over by an annotator, and may be considered errors in the gold standard. For example, the model easily notices that question marks indicate the \textsc{solutionhood} relation, even where these were skipped by annotators in favor of marking WH words instead:

\ex. \textcolor[RGB]{230, 230, 230}{Which} \textcolor[RGB]{230, 230, 230}{previous} \textcolor[RGB]{230, 230, 230}{Virginia} \textcolor[RGB]{230, 230, 230}{Governor(s)} \textcolor[RGB]{230, 230, 230}{do} \textcolor[RGB]{230, 230, 230}{you} \textcolor[RGB]{230, 230, 230}{most} \textcolor[RGB]{230, 230, 230}{admire} \textcolor[RGB]{230, 230, 230}{and} \textcolor[RGB]{230, 230, 230}{why} \textbf{\textcolor[RGB]{12, 12, 12}{?}}  $\xrightarrow[\text{pred:solutionhood}]{\text{gold:solutionhood}}$ \textcolor[RGB]{230, 230, 230}{Thomas} \textcolor[RGB]{230, 230, 230}{Jefferson} \textcolor[RGB]{183, 183, 183}{.} 
 \label{ex:which}

From the model's perspective, the question mark, which scores ${\Delta}_s$=0.79, is the single most important signal, and virtually sufficient for classifying the relation correctly, though it was left out of the gold annotations. The WH word \textit{Which} and the sentence final \textit{why}, by contrast, were noticed by annotators but were are not as unambiguous (the former could be a determiner, and the latter in sentence final position could be part of an embedded clause). In the presence of the question mark, their individual removal has much less impact on the classification decision. Although the model's behavior is sensible and can reveal annotation errors, it also suggests that ${\Delta}_s$ will be blind to auxiliary signals in the presence of very strong, independently sufficient cues.

Using the difference in likelihood of correct relation prediction as a metric also raises the possibility of an opposite concept to signals, which we will refer to as \textit{distractors}. Since ${\Delta}_s$ is a signed measure of difference, it is in fact possible to obtain negative values whenever the removal or masking of a word results in an improvement in the model's ability to predict the relation. In such cases, and especially when the negative value is of a large magnitude, it seems like a reasonable interpretation to say that a word functions as a sort of anti-signal, preventing or complicating the recognition of what might otherwise be a more clear-cut case. Examples \ref{ex:howdo}--\ref{ex:dunno} show some instances of distractors identified by the masking procedure (distractors with ${\Delta}_s<$-0.2 are underlined).

\ex. \underline{\textcolor[RGB]{230, 230, 230}{How}} \underline{\textcolor[RGB]{230, 230, 230}{do}} \textcolor[RGB]{230, 230, 230}{they} \textcolor[RGB]{201, 201, 201}{treat} \textcolor[RGB]{167, 167, 167}{those} \textcolor[RGB]{210, 210, 210}{not} \textcolor[RGB]{190, 190, 190}{like} \textcolor[RGB]{230, 230, 230}{themselves} \textcolor[RGB]{100, 100, 100}{?}  $\xrightarrow[\text{pred:solutionhood}]{\text{gold:preparation}}$ \textcolor[RGB]{52, 52, 52}{then} \textcolor[RGB]{230, 230, 230}{they} \textcolor[RGB]{230, 230, 230}{'re} \textcolor[RGB]{230, 230, 230}{either} \textcolor[RGB]{230, 230, 230}{over-zealous} \textcolor[RGB]{230, 230, 230}{,} \textcolor[RGB]{230, 230, 230}{ignorant} \textcolor[RGB]{230, 230, 230}{of} \textcolor[RGB]{230, 230, 230}{other} \textcolor[RGB]{230, 230, 230}{people} \textcolor[RGB]{230, 230, 230}{or} \textcolor[RGB]{230, 230, 230}{what} \textcolor[RGB]{230, 230, 230}{to} \textcolor[RGB]{230, 230, 230}{avoid} \textcolor[RGB]{230, 230, 230}{those} \textcolor[RGB]{230, 230, 230}{that} \textcolor[RGB]{230, 230, 230}{contradict} \textcolor[RGB]{230, 230, 230}{their} \textcolor[RGB]{230, 230, 230}{fantasy} \textcolor[RGB]{230, 230, 230}{land} \textcolor[RGB]{230, 230, 230}{that} \textcolor[RGB]{220, 220, 220}{caters} \textcolor[RGB]{230, 230, 230}{to} \textcolor[RGB]{230, 230, 230}{them} \textcolor[RGB]{230, 230, 230}{and} \textcolor[RGB]{230, 230, 230}{them} \underline{\textcolor[RGB]{230, 230, 230}{only}} \textcolor[RGB]{230, 230, 230}{.} \label{ex:howdo}

\ex. \textcolor[RGB]{230, 230, 230}{God} \textcolor[RGB]{230, 230, 230}{,} \textcolor[RGB]{230, 230, 230}{I} \textcolor[RGB]{230, 230, 230}{do} \textcolor[RGB]{230, 230, 230}{n't} \textcolor[RGB]{230, 230, 230}{know} \textcolor[RGB]{51, 51, 51}{!}  $\xrightarrow[\text{pred:preparation}]{\text{gold:preparation}}$ \underline{\textcolor[RGB]{230, 230, 230}{but}} \textcolor[RGB]{230, 230, 230}{nobody} \textcolor[RGB]{230, 230, 230}{will} \textcolor[RGB]{230, 230, 230}{go} \textcolor[RGB]{230, 230, 230}{to} \textcolor[RGB]{230, 230, 230}{fight} \textcolor[RGB]{230, 230, 230}{for} \textcolor[RGB]{230, 230, 230}{noses} \textcolor[RGB]{230, 230, 230}{any} \textcolor[RGB]{219, 219, 219}{more} \textcolor[RGB]{169, 169, 169}{.} 

In \ref{ex:howdo}, a rhetorical question trips up the classifier, which predicts the question-answer relation \textsc{solutionhood} instead of \textsc{preparation}. Here the initial WH word \textit{How} and the subsequent auxiliary \textit{do}-support both distract (with ${\Delta}_s$=-0.23 and -0.25) from the \textsc{preparation} relation, which is however being signaled positively by the DM \textit{then} in the nucleus unit. Later on, the adverb \textit{only} is also disruptive (${\Delta}_s$=-0.31), perhaps due to a better association with adversative relations, such as \textsc{contrast}.\label{ex:dunno}

In \ref{ex:dunno}, a preparatory ``God, I don't know!'' is followed up with a nucleus starting with \textit{but}, which typically marks a \textsc{concession} or other adversative relation. In fact, the DM \textit{but} is related to a concessive relation with another EDU (not shown), which the model is not aware of while making the classification for the \textsc{preparation}. Although this example reveals a weakness in the model's inability to consider broader context, it also reveals the difficulty of expecting DMs to fall in line with a strong nuclearity assumption: since units serve multiple functions as satellites and nuclei, signals which aid the recognition of one relation may hinder the recognition of another.

\paragraph{Performance on Signal Types} To better understand the kinds of signals which the model captures better or worse, Table \ref{tab:sig_errtypes} gives a breakdown of performance by signal type and specific signal categories, for categories attested over 20 times (note that the categories are human labels assigned to the corresponding positions -- the system does not predict signal types). To evaluate performance for all types we cannot use recall@1--3, since some sentences contain more than 3 signal tokens, which would lead to recall errors even if the top 3 ranks are correctly identified signals. The scores in the table therefore express how many of the signal tokens belonging to each subtype in the gold annotations are recognized if we allow the system to make as many guesses as there are signal tokens in each EDU pair, plus a tolerance of a maximum of 2 additional tokens (similarly to recall@3). We also note that a single token may be associated with multiple signal types, in which case its identification or omission is counted separately for each type.

\begin{table}[h!tb]
\centering
\begin{tabular}{l|l|r|r}
\textit{type} & \textit{subtype} & \textit{accuracy} & $N_{test}$  \\
\hline
Lexical & Alternate expression & 0.773 & 53 \\
Semantic & Synonymy & 0.727 & 22 \\
Lexical & Indicative word & 0.726 & 340 \\
Graphical & Colon & 0.714 & 21 \\
Morphological & Tense & 0.711 & 59 \\
Semantic & Antonymy & 0.690 & 42 \\
DM & DM & 0.653 & 257 \\
Semantic & Repetition & 0.642 & 350 \\
Numerical & Same count & 0.617 & 34 \\
Semantic + syntactic & Meronymy + subject NP & 0.567 & 37 \\
Semantic + syntactic & Lexical chain + subject NP & 0.566 & 83 \\
Semantic & Lexical chain & 0.549 & 1453 \\
Semantic & Meronymy & 0.545 & 121 \\
Semantic & Indicative word pair & 0.528 & 70 \\
Semantic + syntactic & Repetition + subject NP & 0.505 & 99 \\
Reference & Personal reference & 0.461 & 206 \\
Morphological & Modality & 0.461 & 52 \\
Reference + syntactic & Propositional reference + subject NP & 0.409 & 22 \\
Reference & Demonstrative reference & 0.380 & 84 \\
Syntactic + semantic & Parallel syntactic construction + lexical chain & 0.333 & 99 \\
Textual & Date & 0.318 & 22 \\
Reference & Propositional reference & 0.285 & 21 \\
\end{tabular}
\caption{Anchored token detection accuracy for signal types attested over 20 times.}
\label{tab:sig_errtypes}
\end{table}

Three of the top four categories which the model performs best for are, perhaps unsurprisingly, the most lexical ones: \textit{alternate expression} captures non-DM phrases such as \textit{I mean} (for \textsc{elaboration}), or \textit{the problem is} (for \textsc{concession}), and \textit{indicative word} includes lexical items such as imperative \textit{see} (consistently marking \textsc{evidence} in references within academic articles) or evaluative adjectives such as \textit{interesting} for \textsc{evaluation}. The good performance of the category \textit{colon} captures the model's recognition of colons as important punctuation, primarily predicting \textsc{preparation}. The only case of a `relational' category, requiring attention to two separate positions in the input, which also fares well is \textit{synonymy}, though this is often based on flagging only one of two items annotated as synonymous, and is based on rather few examples. We can find only one example, \ref{ex:decrease}, where both sides of a pair of similar words is actually noticed, which both belong to the same stem (\textit{decline/declining}):

\ex. \textcolor[RGB]{230, 230, 230}{The} \textcolor[RGB]{230, 230, 230}{report} \textcolor[RGB]{209, 209, 209}{says} \textcolor[RGB]{213, 213, 213}{the} \textcolor[RGB]{172, 172, 172}{decline} \textcolor[RGB]{220, 220, 220}{in} \textcolor[RGB]{228, 228, 228}{iodine} \textcolor[RGB]{230, 230, 230}{intake} \textcolor[RGB]{215, 215, 215}{appears} \textcolor[RGB]{230, 230, 230}{to} \textcolor[RGB]{230, 230, 230}{be} \textcolor[RGB]{230, 230, 230}{due} \textcolor[RGB]{230, 230, 230}{to} \textcolor[RGB]{230, 230, 230}{changes} \textcolor[RGB]{230, 230, 230}{in} \textcolor[RGB]{230, 230, 230}{the} \textcolor[RGB]{230, 230, 230}{dairy} \textcolor[RGB]{230, 230, 230}{industry} \textcolor[RGB]{230, 230, 230}{,} \textcolor[RGB]{230, 230, 230}{where} \textcolor[RGB]{230, 230, 230}{chlorine-containing} \textcolor[RGB]{230, 230, 230}{sanitisers} \textcolor[RGB]{226, 226, 226}{have} \textcolor[RGB]{230, 230, 230}{replaced} \textcolor[RGB]{230, 230, 230}{iodine-containing} \textcolor[RGB]{230, 230, 230}{sanitisers} \textcolor[RGB]{230, 230, 230}{.}  $\xleftarrow[\text{pred:background}]{\text{gold:justify}}$ \textcolor[RGB]{193, 193, 193}{Iodine} \textcolor[RGB]{230, 230, 230}{released} \textcolor[RGB]{230, 230, 230}{from} \textcolor[RGB]{230, 230, 230}{these} \textcolor[RGB]{230, 230, 230}{chemicals} \textcolor[RGB]{230, 230, 230}{into} \textcolor[RGB]{216, 216, 216}{milk} \textcolor[RGB]{230, 230, 230}{has} \textcolor[RGB]{230, 230, 230}{been} \textcolor[RGB]{230, 230, 230}{the} \textcolor[RGB]{230, 230, 230}{major} \textcolor[RGB]{230, 230, 230}{source} \textcolor[RGB]{230, 230, 230}{of} \textcolor[RGB]{226, 226, 226}{dietary} \textcolor[RGB]{206, 206, 206}{iodine} \textcolor[RGB]{230, 230, 230}{in} \textcolor[RGB]{230, 230, 230}{Australia} \textcolor[RGB]{230, 230, 230}{for} \textcolor[RGB]{230, 230, 230}{at} \textcolor[RGB]{230, 230, 230}{least} \textcolor[RGB]{230, 230, 230}{four} \textcolor[RGB]{230, 230, 230}{decades} \textcolor[RGB]{202, 202, 202}{,} \textcolor[RGB]{153, 153, 153}{but} \textcolor[RGB]{230, 230, 230}{is} \textcolor[RGB]{230, 230, 230}{now} \textbf{\textcolor[RGB]{63, 63, 63}{declining}} \textcolor[RGB]{79, 79, 79}{.} \label{ex:decrease}

We note that our evaluation is actually rather harsh towards the model, since in multiword expressions, often only one central word  is flagged by ${\Delta}_s$ (e.g.~\textit{problem} in \textit{``the problem is''}), while the model is penalized in Table \ref{tab:sig_errtypes} for each token that is not recognized (i.e.~\textit{the} and \textit{is}, which were all flagged by a human annotator as signals in the data). 

Interestingly, the model fares rather well in identifying morphological tense cues, even though these are marked by both inflected lexical verbs and semantically poor auxiliaries (e.g.~past perfect auxiliary \textit{had} marking \textsc{background}); but modality cues (especially \textit{can} or \textit{could} for \textsc{evaluation}) are less successfully identified, suggesting they are either more ambiguous, or mainly relevant in the presence of evaluative content words which out-score them.

Other relational categories from the middle of the table which ostensibly require matching pairs of words, such as \textit{repetition}, \textit{meronymy}, or \textit{personal reference} (coreference) are mainly captured by the model when a single item is a sufficiently powerful cue, often ignoring the other half of the signal, as shown in \ref{ex:its}.

\ex. \textcolor[RGB]{230, 230, 230}{On} \textcolor[RGB]{230, 230, 230}{a} \textcolor[RGB]{230, 230, 230}{new} \textcolor[RGB]{230, 230, 230}{website} \textcolor[RGB]{230, 230, 230}{,} \textcolor[RGB]{230, 230, 230}{"} \textcolor[RGB]{230, 230, 230}{The} \textcolor[RGB]{230, 230, 230}{Internet} \textcolor[RGB]{230, 230, 230}{Explorer} \textcolor[RGB]{230, 230, 230}{6} \textcolor[RGB]{230, 230, 230}{Countdown} \textcolor[RGB]{230, 230, 230}{"} \textcolor[RGB]{230, 230, 230}{,} \textcolor[RGB]{230, 230, 230}{Microsoft} \textcolor[RGB]{230, 230, 230}{has} \textcolor[RGB]{230, 230, 230}{launched} \textcolor[RGB]{230, 230, 230}{an} \textcolor[RGB]{230, 230, 230}{aggressive} \textcolor[RGB]{230, 230, 230}{campaign} \textcolor[RGB]{230, 230, 230}{to} \textcolor[RGB]{230, 230, 230}{persuade} \textcolor[RGB]{230, 230, 230}{users} \textcolor[RGB]{230, 230, 230}{to} \textcolor[RGB]{230, 230, 230}{stop} \textcolor[RGB]{171, 171, 171}{using} \textcolor[RGB]{133, 133, 133}{IE6}  $\xleftarrow[\text{pred:elaboration}]{\text{gold:elaboration}}$ \textbf{\textcolor[RGB]{56, 56, 56}{Its}} \textcolor[RGB]{197, 197, 197}{goal} \textcolor[RGB]{167, 167, 167}{is} \textcolor[RGB]{230, 230, 230}{to} \textcolor[RGB]{230, 230, 230}{decrease} \textcolor[RGB]{230, 230, 230}{IE6} \textcolor[RGB]{230, 230, 230}{users} \textcolor[RGB]{230, 230, 230}{to} \textcolor[RGB]{230, 230, 230}{less} \textcolor[RGB]{230, 230, 230}{than} \textcolor[RGB]{230, 230, 230}{one} \textcolor[RGB]{124, 124, 124}{percent} \textcolor[RGB]{229, 229, 229}{.} \label{ex:its}

Here the model has learned that an initial possessive pronoun, perhaps in the context of a subject NP in a copula sentence (note the shading of the following \textit{is}) is an indicator of an \textsc{elaboration} relation, even though there is no indication that the model has noticed which word is the antecedent. Similarly for the \textit{count} category, the model only learns to notice the possible importance of some numbers, but is not actually aware of whether they are identical (e.g.~for \textsc{restatement}) or different (e.g.~in \textsc{contrast}).\footnote{We note that while the original Signalling Corpus only used a \textit{Same count} specific signal category, the data set we are working with uses a similar specific signal type for any numerical signals, including \textit{first(ly)} and \textit{second(ly)}, which the model is particularly successful in learning.}

Finally, some categories are actually recognized fairly reliably, but are penalized by the same partial substring issue identified above: Date expressions are consistently flagged as indicators of \textsc{circumstance}, but often a single word, such as a weekday in \ref{ex:wednesday}, is dominant, while the model is penalized for not scoring other words as highly (including commas within dates, which are marked as part of the signal token span in the gold standard, but whose removal does not degrade prediction accuracy). In this case it seems fair to say that the model has successfully recognized the date signal of `Wednesday April 13', yet it loses points for missing two instances of `,', and the `2011', which is no longer necessary for recognizing that this is a date.

\ex. \textcolor[RGB]{230, 230, 230}{NASA} \textcolor[RGB]{230, 230, 230}{celebrates} \textcolor[RGB]{230, 230, 230}{30th} \textcolor[RGB]{230, 230, 230}{anniversary} \textcolor[RGB]{230, 230, 230}{of} \textcolor[RGB]{230, 230, 230}{first} \textcolor[RGB]{230, 230, 230}{shuttle} \textcolor[RGB]{230, 230, 230}{launch} \textcolor[RGB]{230, 230, 230}{;}  $\xleftarrow[\text{pred:circumstance}]{\text{gold:circumstance}}$ \textbf{\textcolor[RGB]{11, 11, 11}{Wednesday}} \textcolor[RGB]{186, 186, 186}{,} \textcolor[RGB]{115, 115, 115}{April} \textcolor[RGB]{153, 153, 153}{13} \textcolor[RGB]{219, 219, 219}{,} \textcolor[RGB]{230, 230, 230}{2011} \label{ex:wednesday}

\section{Discussion} \label{discussion}

This paper has used a corpus annotated for discourse relation signals within the framework of the RST Signalling Corpus (\citealt{DasTaboada2017}) and extended with anchored signal annotations (\citealt{liu-2019-beyond}) to develop a taxonomy of unrestricted and hierarchically aware discourse signal positions, as well as a data-driven neural network model to explore distantly supervised signal word extraction. The results shed light on the distribution of signal categories from the RST-SC taxonomy in terms of associated word forms, and show the promise of neural models with contextual embeddings for the extraction of context dependent and gradient discourse signal detection in individual texts. The metric developed for the evaluation, $\Delta _s$, allows us to assess the relative importance of signal words for automatic relation classification, and reveal observations for further study, as well as shortcomings which point to the need to develop richer feature representations and system architectures in future work.

The model presented in the previous sections is clearly incomplete in both its classification accuracy and its ability to recognize the same signals that humans do. However, given the fact that it is trained entirely without access to discourse signal annotations and is unaware of any of the guidelines used to create the gold standard that it is evaluated on, its performance may be considered surprisingly good. As an approach to extracting discourse signals in a data-driven way, similar to frequentist methods or association measures used in previous work, we suggest that this model forms a more fine grained tool, capable of taking context into consideration and delivering scores for each instance of a signal candidate, rather than resulting in a table of undifferentiated signal word types. 

Additionally, although we consider human signal annotations to be the gold standard in identifying the presence of relevant cues, the ${\Delta}_s$ metric gives new insights into signaling which cannot be approached using manual signaling annotations. Firstly, the quantitative nature of the metric allows us to rank signaling strength in a way that humans have not to date been able to apply: using ${\Delta}_s$, we can say which instances of which signals are evaluated as stronger, by how much, and which words within a multi-word signal instance are the most important (e.g.~weekdays in dates are important, the commas are not). Secondly, the potential for negative values of the metric opens the door to the study of negative signals, or `distractors', which we have only touched upon briefly in this paper. And finally, we consider the availability of multiple measurements for a single DM or other discourse signal to be a potentially very interesting window into the relative ambiguity of different signaling devices (cf.~\citealt{TorabiAsrDemberg2013}) and for research on the contexts in which such ambiguity results. 

\begin{figure*}[h!tb]
\centering
\includegraphics[width=120mm]{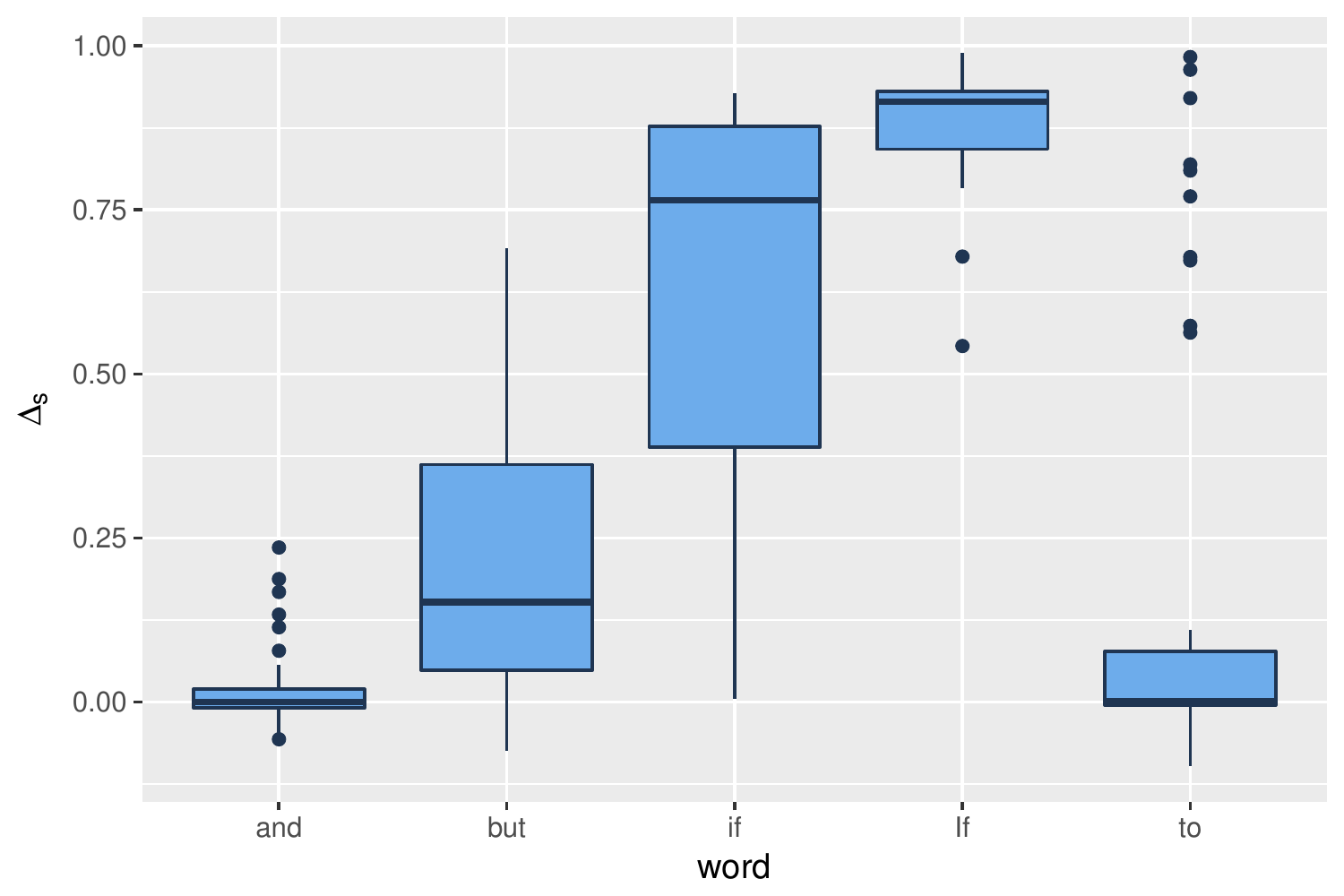}
\caption{Boxplots for all ${\Delta}_s$ values of several signal token types across the test corpus.}
\label{fig:ambig_signals}
\par\vspace{-5pt}\par
\end{figure*}

To see how ambiguity is reflected in multiple measurements of  ${\Delta}_s$, we can consider Figure \ref{fig:ambig_signals}. The figure shows boxplots for multiple instances of the same signal tokens. We can see that words like \textit{and} are usually not strong signals, with the entire interquartile range scoring less than 0.02, i.e.~aiding relation classification by less than 2\%, with some values dipping into the negative region (i.e.~cases functioning as distractors). However, some outliers are also present, reaching almost as high as 0.25 -- these are likely to be coordinating predicates, which may signal relations such as \textsc{sequence} or \textsc{joint}. A word such as \textit{but} is more important overall, with the box far above \textit{and}, but still covering a wide range of values: these can correspond to more or less ambiguous cases of \textit{but}, but also to cases in which the word is more or less irreplaceable as a signal. In the presence of multiple signals for the same relation, the presence of \textit{but} should be less important. We can also see that \textit{but} can be a distractor with negative values, as we saw in example \ref{ex:dunno} above. As far as we are aware, this is the first empirical corpus-based evidence giving a quantitative confirmation to the intuition that `but' in context is significantly less ambiguous as a discourse marker than `and'; the overlap in their bar plots indicate that they can be similarly ambiguous or even distracting in some cases, but the difference in interquartile ranges makes it clear that these are exceptions.

For less ambiguous DMs, such as \textit{if}, we can also see a contrast between lower and upper case instances: upper case \textit{If} is almost always a marker of \textsc{condition}, but the lower case \textit{if} is sometimes part of an embedded object clause, which is not segmented in the corpus and does not mark a conditional relation (e.g.~``they wanted to see if...''). For the word \textit{to}, the figure suggests a strongly bimodal distribution, with a core population of (primarily prepositional) discourse-irrelevant \textit{to}, and a substantial number of outliers above a large gap, representing \textit{to} in infinitival \textsc{purpose} clauses (though not all \textit{to} infinitives mark such clauses, as in adnominal ``a chance to go'', which the model is usually able to distinguish in context). In other words, our model can not only disambiguate ambiguous strings into grammatical categories, but also rank members of the same category by importance in context, as evidenced by its ability to correctly classify high frequency items like `to' or `and' as true positives. A frequentist approach would not only lack this ability -- it would miss such items altogether, due to its overall high string frequency and low specificity.

Beyond what the results can tell us about discourse signals in this particular corpus, the fact that the neural model is sensitive to mutual redundancy of signals raises interesting theoretical questions about what human annotators are doing when they characterize multiple features of a discourse unit as signals. If it is already evident from the presence of a conventional DM that some relation applies, are other, less explicit signals which might be relied on in the absence of the DM, equally `there'? Do we need a concept of primary and auxiliary signals, or graded signaling strength, in the way that a metric such as ${\Delta}_s$ suggests?

Another open question relates to the postulation of distractors as an opposite concept to discourse relation signals. While we have not tested this so far, it is interesting to ask to what extent human analysts are aware of distractors, whether we could form annotation guidelines to recognize them, and how humans weigh the value of signals and potential distractors in extrapolating intended discourse relations. It seems likely that distractors affecting humans may be found in cases of misunderstanding or ambiguity of discourse relations (see also \citealt{Cunha2013}). 

Finally, the error analysis for signal detection complements the otherwise opaque relation classification results in Table \ref{tab:model_performance} in showing some of the missing sources of information that our model would need in order to work better. We have seen that relational information, such as identifying not just the presence of a pronoun but also its antecedent, or both sides of lexical semantic relations such as synonymy, meronymy or antonymy, as well as comparing count information, are still unavailable to the classifier -- if they were being used, then ${\Delta}_s$ would reflect the effects of their removal, but this is largely not the case. This suggests that, in the absence of vastly larger discourse annotated corpora, discourse relation recognition may require the construction of either features, architectures, or both, which can harness abstract relational information of this nature beyond the memorization of specific pairs of words (or regions of vector space with similar words) that are already attested in the limited training data. In this vein, \cite{pitler-etal-2009-automatic} conducted a series of experiments on automatic sense prediction for four top-level implicit discourse relations within the PDTB framework, which also suggested benefits for using linguistically-informed features such as verb information, polarity tags, context, lexical items (e.g.~first and last words of the arguments; first three words in the sentence) etc. The model architecture and input data are also in need of improvements, as the current architecture can only be expected to identify endocentric signals. The substantial amount of exocentric signaling cases is in itself an interesting finding, as it suggests that relation classification from head EDU pairs may ultimately have a natural ceiling that is considerably below what could be inferred from looking at larger contexts. We predict that as we add more features to the model and improve its architecture in ways that allow it to recognize the kinds of signals that humans do, classification accuracy will increase; and conversely, as classification accuracy rises, measurements based on ${\Delta}_s$ will overlap increasingly with human annotations of anchored signals.

In sum, we believe that there is room for much research on what relation classification models should look like, and how they can represent the kinds of information found in non-trivial signals. The results of this line of work can therefore benefit NLP systems targeting discourse relations by suggesting locations within the text which systems should attend to in one way or another. Moreover, we think that using  distant-supervised techniques for learning discourse relations (e.g.~\citealt{huber-carenini-2019-predicting}) is promising in the development of discourse models using the proposed dataset. We hope to see further analyses benefit from this work and the application of metrics such as ${\Delta}_s$ to other datasets, within more complex models, and using additional features to capture such information. We also hope to see applications of discourse relations such as machine comprehension \citep{narasimhan-barzilay-2015-machine} and sentiment analysis \citep{huber-carenini-2019-predicting} etc.~benefit from the proposed model architecture as well as the dataset.


\bibliography{bibl.bib}

\begin{thebibliography}{56}
\providecommand{\natexlab}[1]{#1}
\providecommand{\url}[1]{\texttt{#1}}
\expandafter\ifx\csname urlstyle\endcsname\relax
  \providecommand{\doi}[1]{doi: #1}\else
  \providecommand{\doi}{doi: \begingroup \urlstyle{rm}\Url}\fi

\bibitem[Akbik et~al.(2019)Akbik, Bergmann, and
  Vollgraf]{AkbikBergmannVollgraf2019}
Alan Akbik, Tanja Bergmann, and Roland Vollgraf.
\newblock Pooled contextualized embeddings for named entity recognition.
\newblock In \emph{Proceedings of NAACL 2019}, pages 724--728, Minneapolis, MN,
  2019.

\bibitem[Benamara et~al.(2017)Benamara, Taboada, and
  Mathieu]{BenamaraTaboadaMathieu2017}
Farah Benamara, Maite Taboada, and Yannick Mathieu.
\newblock Evaluative language beyond bags of words: Linguistic insights and
  computational applications.
\newblock \emph{Computational Linguistics}, 43\penalty0 (1):\penalty0 201--264,
  2017.

\bibitem[Carlson et~al.(2003)Carlson, Marcu, and Okurowski]{CarlsonEtAl2003}
Lynn Carlson, Daniel Marcu, and Mary~Ellen Okurowski.
\newblock Building a discourse-tagged corpus in the framework of {Rhetorical
  Structure Theory}.
\newblock In \emph{Current and New Directions in Discourse and Dialogue}, Text,
  Speech and Language Technology 22, pages 85--112. Kluwer, Dordrecht, 2003.

\bibitem[Casalicchio et~al.(2019)Casalicchio, Molnar, and
  Bischl]{CasalicchioMolnarBischl2019}
Giuseppe Casalicchio, Christoph Molnar, and Bernd Bischl.
\newblock Visualizing the feature importance for black box models.
\newblock In Michele Berlingerio, Francesco Bonchi, Thomas G\"{a}rtner, Neil
  Hurley, and Georgiana Ifrim, editors, \emph{Machine Learning and Knowledge
  Discovery in Databases. ECML PKDD 2018}, Lecture Notes in Computer Science
  11051. Springer, Cham, 2019.

\bibitem[da~Cunha(2013)]{Cunha2013}
Iria da~Cunha.
\newblock A symbolic corpus-based approach to detect and solve the ambiguity of
  discourse markers.
\newblock \emph{Research in Computing Science}, 70:\penalty0 95--106, 2013.

\bibitem[Das and Taboada(2017)]{DasTaboada2017}
Debopam Das and Maite Taboada.
\newblock Signalling of coherence relations in discourse, beyond discourse
  markers.
\newblock \emph{Discourse Processes}, pages 1--29, 2017.

\bibitem[Das and Taboada(2018)]{das2018rst}
Debopam Das and Maite Taboada.
\newblock R{ST} {S}ignalling {C}orpus: A corpus of signals of coherence
  relations.
\newblock \emph{Language Resources and Evaluation}, 52\penalty0 (1):\penalty0
  149--184, 2018.

\bibitem[Das et~al.(2019)Das, Taboada, and
  McFetridge]{DasTaboadaMcFetridge2019}
Debopam Das, Maite Taboada, and Paul McFetridge.
\newblock {RST Signalling Corpus. LDC2015T10}, 2019.

\bibitem[Duque(2014)]{Duque2014}
Eldadio Duque.
\newblock Signaling causal coherence relations.
\newblock \emph{Discourse Studies}, 16\penalty0 (1):\penalty0 25--46, 2014.

\bibitem[Francis(1994)]{Francis1994}
Gill Francis.
\newblock Labelling discourse: An aspect of nominal-group lexical cohesion.
\newblock In Malcolm Coulthard, editor, \emph{Advances in Written Text
  Analysis}, pages 83--101. Routledge, 1994.

\bibitem[Fraser(1999)]{Fraser1999}
Bruce Fraser.
\newblock What are discourse markers?
\newblock \emph{Journal of Pragmatics}, 31:\penalty0 931--952, 1999.

\bibitem[Gardner et~al.(2018)Gardner, Grus, Neumann, Tafjord, Dasigi, Liu,
  Peters, Schmitz, and Zettlemoyer]{GardnerGrusNeumannEtAl2018}
Matt Gardner, Joel Grus, Mark Neumann, Oyvind Tafjord, Pradeep Dasigi,
  Nelson~F. Liu, Matthew Peters, Michael Schmitz, and Luke~S. Zettlemoyer.
\newblock {AllenNLP}: A deep semantic natural language processing platform.
\newblock In \emph{Proceedings of ACL 2018}, pages 1--6, Melbourne, 2018.

\bibitem[Gessler et~al.(2019)Gessler, Liu, and
  Zeldes]{gessler-etal-2019-discourse}
Luke Gessler, Yang Liu, and Amir Zeldes.
\newblock A discourse signal annotation system for {RST} trees.
\newblock In \emph{Proceedings of the Workshop on Discourse Relation Parsing
  and Treebanking (DISRPT 2019)}, pages 56--61, Minneapolis, MN, 2019.

\bibitem[Ghaeini et~al.(2018)Ghaeini, Fern, and
  Tadepalli]{GhaeiniFernTadepalli2018}
Reza Ghaeini, Xiaoli~Z. Fern, and Prasad Tadepalli.
\newblock Interpreting recurrent and attention-based neural models: A case
  study on natural language inference.
\newblock In \emph{Proceedings of EMNLP 2018}, pages 4952--4957, Brussels,
  2018.

\bibitem[Hovy and Maier(1993)]{HovyMaier1993}
Eduard~H. Hovy and Elisabeth Maier.
\newblock Parsimonious or profligate: How many and which discourse structure
  relations?
\newblock Technical report, Information Sciences Institute, USC, 1993.

\bibitem[Huber and Carenini(2019)]{huber-carenini-2019-predicting}
Patrick Huber and Giuseppe Carenini.
\newblock Predicting discourse structure using distant supervision from
  sentiment.
\newblock In \emph{Proceedings of the 2019 Conference on Empirical Methods in
  Natural Language Processing and the 9th International Joint Conference on
  Natural Language Processing (EMNLP-IJCNLP)}, pages 2306--2316, Hong Kong,
  China, November 2019.

\bibitem[Knott(1996)]{Knott1996}
Alistair Knott.
\newblock \emph{A Data-Driven Methodology for Motivating a Set of Coherence
  Relations}.
\newblock PhD thesis, University of Edinburgh, 1996.

\bibitem[Knott and Dale(1994)]{KnottDale1994}
Alistair Knott and Robert Dale.
\newblock Using linguistic phenomena to motivate a set of coherence relations.
\newblock \emph{Discourse Processes}, 18\penalty0 (1):\penalty0 35--52, 1994.

\bibitem[Knott and Sanders(1998)]{knott1998classification}
Alistair Knott and Ted Sanders.
\newblock The classification of coherence relations and their linguistic
  markers: {A}n exploration of two languages.
\newblock \emph{Journal of Pragmatics}, 30\penalty0 (2):\penalty0 135--175,
  1998.

\bibitem[Leech et~al.(2003)Leech, McEnery, and Weisser]{LeechEtAl2003}
Geoffrey Leech, Tony McEnery, and Martin Weisser.
\newblock {SPAAC} speech-act annotation scheme.
\newblock Technical report, Lancaster University, 2003.

\bibitem[Liu(2019)]{liu-2019-beyond}
Yang Liu.
\newblock Beyond the {Wall Street Journal}: Anchoring and comparing discourse
  signals across genres.
\newblock In \emph{Proceedings of the Workshop on Discourse Relation Parsing
  and Treebanking (DISRPT 2019)}, pages 72--81, Minneapolis, MN, 2019.

\bibitem[Liu and Zeldes(2019)]{liu2019discourse}
Yang Liu and Amir Zeldes.
\newblock Discourse relations and signaling information: Anchoring discourse
  signals in {RST-DT}.
\newblock \emph{Proceedings of the Society for Computation in Linguistics},
  2\penalty0 (1):\penalty0 314--317, 2019.

\bibitem[Mann and Thompson(1988)]{MannThompson1988}
William~C. Mann and Sandra~A. Thompson.
\newblock {Rhetorical Structure Theory}: Toward a functional theory of text
  organization.
\newblock \emph{Text}, 8\penalty0 (3):\penalty0 243--281, 1988.

\bibitem[Marcu(1996)]{Marcu1996}
Daniel Marcu.
\newblock Building up rhetorical structure trees.
\newblock In \emph{Proceedings of AAAI-96}, pages 1069--1074, Portland, OR,
  1996.

\bibitem[Marcus et~al.(1993)Marcus, Santorini, and
  Marcinkiewicz]{MarcusSantoriniMarcinkiewicz1993}
Mitchell~P. Marcus, Beatrice Santorini, and Mary~Ann Marcinkiewicz.
\newblock Building a large annotated corpus of {E}nglish: The {P}enn
  {T}reebank.
\newblock \emph{Special Issue on Using Large Corpora, Computational
  Linguistics}, 19\penalty0 (2):\penalty0 313--330, 1993.

\bibitem[M\'{i}rovsk\'{y} et~al.(2017)M\'{i}rovsk\'{y}, Synkov\'{a},
  Rysov\'{a}, and Pol\'{a}kov\'{a}]{MirovskySynkovaRysovaEtAl2017}
Ji\v{r}\'{i} M\'{i}rovsk\'{y}, Pavl\'{i}na Synkov\'{a}, Magdal\'{e}na
  Rysov\'{a}, and Lucie Pol\'{a}kov\'{a}.
\newblock {CzeDLex} – a lexicon of {C}zech discourse connectives.
\newblock \emph{The Prague Bulletin of Mathematical Linguistics}, 109:\penalty0
  61--91, 2017.

\bibitem[Morey et~al.(2017)Morey, Muller, and Asher]{MoreyMullerAsher2017}
Mathieu Morey, Philippe Muller, and Nicholas Asher.
\newblock How much progress have we made on {RST} discourse parsing? {A}
  replication study of recent results on the {RST-DT}.
\newblock In \emph{Proceedings of EMNLP 2017}, pages 1319--1324, Copenhagen,
  Denmark, 2017.

\bibitem[Muller et~al.(2019)Muller, Braud, and Morey]{MullerBraudMorey2019}
Philippe Muller, Chlo\'{e} Braud, and Mathieu Morey.
\newblock {ToNy}: Contextual embeddings for accurate multilingual discourse
  segmentation of full documents.
\newblock In \emph{Proceedings of Discourse Relation Treebanking and Parsing
  (DISRPT 2019)}, pages 115--124, Minneapolis, MN, 2019.

\bibitem[Narasimhan and Barzilay(2015)]{narasimhan-barzilay-2015-machine}
Karthik Narasimhan and Regina Barzilay.
\newblock Machine comprehension with discourse relations.
\newblock In \emph{Proceedings of the 53rd Annual Meeting of the Association
  for Computational Linguistics and the 7th International Joint Conference on
  Natural Language Processing (ACL/IJCNLP 2015)}, pages 1253--1262, Beijing,
  China, 2015.

\bibitem[Nivre et~al.(2017)Nivre, \v{Z}eljko Agi{\'c}, Ahrenberg, Aranzabe,
  Asahara, Atutxa, Ballesteros, Bauer, Bengoetxea, Bhat, Bick, Bosco, Bouma,
  Bowman, Candito, Eryi{\u g}it, Celano, Chalub, Choi, \c{C}a\u{g}r{\i}
  \c{C}\"{o}ltekin, Connor, Davidson, de~Marneffe, de~Paiva, de~Ilarraza,
  Dobrovoljc, Dozat, Droganova, Dwivedi, Eli, Erjavec, Farkas, Foster, Freitas,
  Gajdo{\v s}ov{\'a}, Galbraith, Garcia, Ginter, Goenaga, Gojenola,
  G{\"o}k{\i}rmak, Goldberg, Guinovart, Saavedra, Grioni, Gr{\=u}z{\={\i}}tis,
  Guillaume, Habash, Haji{\v c}, M{\~y}, Haug, Hladk{\'a}, Hohle, Ion, Irimia,
  Johannsen, J{\o}rgensen, Ka{\c s}{\i}kara, Kanayama, Kanerva, Kotsyba, Krek,
  Laippala, H{\`{\^o}}ng, Lenci, Ljube{\v s}i{\'c}, Lyashevskaya, Lynn,
  Makazhanov, Manning, M{\u a}r{\u a}nduc, Mare{\v c}ek, Alonso, Martins, Ma{\v
  s}ek, Matsumoto, {McDonald}, Missil{\"a}, Mititelu, Miyao, Montemagni, More,
  Mori, Moskalevskyi, Muischnek, Mustafina, M{\"u}{\"u}risep, Th{\d i}, Minh,
  Nikolaev, Nurmi, Ojala, Osenova, {{\O}}vrelid, Pascual, Passarotti, Perez,
  Perrier, Petrov, Piitulainen, Plank, Popel, Pretkalni{\c n}a, Prokopidis,
  Puolakainen, Pyysalo, Rademaker, Ramasamy, Real, Rituma, Rosa, Saleh,
  Sanguinetti, Saul{\={\i}}te, Schuster, Seddah, Seeker, Seraji, Shakurova,
  Shen, Sichinava, Silveira, Simi, Simionescu, Simk{\'o}, \v{S}imkov{\'a},
  Simov, Smith, Suhr, Sulubacak, Sz{\'a}nt{\'o}, Taji, Tanaka, Tsarfaty, Tyers,
  Uematsu, Uria, van Noord, Varga, Vincze, Washington, {\\v Z}abokrtsk{\'y},
  Zeldes, Zeman, and Zhu]{NivreEtAl2017}
Joakim Nivre, \v{Z}eljko Agi{\'c}, Lars Ahrenberg, Maria~Jesus Aranzabe,
  Masayuki Asahara, Aitziber Atutxa, Miguel Ballesteros, John Bauer, Kepa
  Bengoetxea, Riyaz~Ahmad Bhat, Eckhard Bick, Cristina Bosco, Gosse Bouma, Sam
  Bowman, Marie Candito, G{\"u}l{\c s}en~Cebiro{\u g}lu Eryi{\u g}it, Giuseppe
  G.~A. Celano, Fabricio Chalub, Jinho Choi, \c{C}a\u{g}r{\i} \c{C}\"{o}ltekin,
  Miriam Connor, Elizabeth Davidson, Marie-Catherine de~Marneffe, Valeria
  de~Paiva, Arantza~Diaz de~Ilarraza, Kaja Dobrovoljc, Timothy Dozat, Kira
  Droganova, Puneet Dwivedi, Marhaba Eli, Toma{\v z} Erjavec, Rich{\'a}rd
  Farkas, Jennifer Foster, Cl{\'a}udia Freitas, Katar{\'{\i}}na Gajdo{\v
  s}ov{\'a}, Daniel Galbraith, Marcos Garcia, Filip Ginter, Iakes Goenaga,
  Koldo Gojenola, Memduh G{\"o}k{\i}rmak, Yoav Goldberg, Xavier~G{\'o}mez
  Guinovart, Berta~Gonz{\'a}les Saavedra, Matias Grioni, Normunds
  Gr{\=u}z{\={\i}}tis, Bruno Guillaume, Nizar Habash, Jan Haji{\v c},
  Linh~H{\`a} M{\~y}, Dag Haug, Barbora Hladk{\'a}, Petter Hohle, Radu Ion,
  Elena Irimia, Anders Johannsen, Fredrik J{\o}rgensen, H{\"u}ner Ka{\c
  s}{\i}kara, Hiroshi Kanayama, Jenna Kanerva, Natalia Kotsyba, Simon Krek,
  Veronika Laippala, L{\^e}~H{\`{\^o}}ng, Alessandro Lenci, Nikola Ljube{\v
  s}i{\'c}, Olga Lyashevskaya, Teresa Lynn, Aibek Makazhanov, Christopher
  Manning, C{\u a}t{\u a}lina M{\u a}r{\u a}nduc, David Mare{\v c}ek,
  H{\'e}ctor~Mart{\'{\i}}nez Alonso, Andr{\'e} Martins, Jan Ma{\v s}ek, Yuji
  Matsumoto, Ryan {McDonald}, Anna Missil{\"a}, Verginica Mititelu, Yusuke
  Miyao, Simonetta Montemagni, Amir More, Shunsuke Mori, Bohdan Moskalevskyi,
  Kadri Muischnek, Nina Mustafina, Kaili M{\"u}{\"u}risep, Luong~Nguy{\~{\^e}}n
  Th{\d i}, Huy{\`{\^e}}n Nguy{\~{\^e}}n~Th{\d i} Minh, Vitaly Nikolaev, Hanna
  Nurmi, Stina Ojala, Petya Osenova, Lilja {{\O}}vrelid, Elena Pascual, Marco
  Passarotti, Cenel-Augusto Perez, Guy Perrier, Slav Petrov, Jussi Piitulainen,
  Barbara Plank, Martin Popel, Lauma Pretkalni{\c n}a, Prokopis Prokopidis,
  Tiina Puolakainen, Sampo Pyysalo, Alexandre Rademaker, Loganathan Ramasamy,
  Livy Real, Laura Rituma, Rudolf Rosa, Shadi Saleh, Manuela Sanguinetti, Baiba
  Saul{\={\i}}te, Sebastian Schuster, Djam{\'e} Seddah, Wolfgang Seeker, Mojgan
  Seraji, Lena Shakurova, Mo~Shen, Dmitry Sichinava, Natalia Silveira, Maria
  Simi, Radu Simionescu, Katalin Simk{\'o}, M{\'a}ria \v{S}imkov{\'a}, Kiril
  Simov, Aaron Smith, Alane Suhr, Umut Sulubacak, Zsolt Sz{\'a}nt{\'o}, Dima
  Taji, Takaaki Tanaka, Reut Tsarfaty, Francis Tyers, Sumire Uematsu, Larraitz
  Uria, Gertjan van Noord, Viktor Varga, Veronika Vincze, Jonathan~North
  Washington, Zden{\v e}k {\\v Z}abokrtsk{\'y}, Amir Zeldes, Daniel Zeman, and
  Hanzhi Zhu.
\newblock Universal dependencies 2.0.
\newblock Technical report, {LINDAT}/{CLARIN} digital library at the Institute
  of Formal and Applied Linguistics ({\'{U}FAL}), Faculty of Mathematics and
  Physics, Charles University, 2017.
\newblock URL \url{http://hdl.handle.net/11234/1-1983}.

\bibitem[Pennington et~al.(2014)Pennington, Socher, and
  Manning]{PenningtonSocherManning2014}
Jeffrey Pennington, Richard Socher, and Christopher~D. Manning.
\newblock Glo{V}e: Global vectors for word representation.
\newblock In \emph{Proceedings of EMNLP 2014}, pages 1532--1543, Doha, Qatar,
  2014.

\bibitem[Peters et~al.(2018)Peters, Neumann, Iyyer, Gardner, Clark, Lee, and
  Zettlemoyer]{PetersNeumannIyyerEtAl2018}
Matthew~E. Peters, Mark Neumann, Mohit Iyyer, Matt Gardner, Christopher Clark,
  Kenton Lee, and Luke Zettlemoyer.
\newblock Deep contextualized word representations.
\newblock In \emph{Proceedings of NAACL 2018}, pages 2227--2237, New Orleans,
  LA, 2018.

\bibitem[Pitler et~al.(2009)Pitler, Louis, and
  Nenkova]{pitler-etal-2009-automatic}
Emily Pitler, Annie Louis, and Ani Nenkova.
\newblock Automatic sense prediction for implicit discourse relations in text.
\newblock In \emph{Proceedings of the Joint Conference of the 47th Annual
  Meeting of the {ACL} and the 4th International Joint Conference on Natural
  Language Processing of the {AFNLP}}, pages 683--691, Suntec, Singapore, 2009.

\bibitem[Poesio et~al.(2002)Poesio, {Di Eugenio}, and
  Keohane]{PoesioEugenioKeohane2002}
Massimo Poesio, Barbara {Di Eugenio}, and Gerard Keohane.
\newblock Discourse structure and anaphora: An empirical study.
\newblock NLE Technical Note TN-02-02, University of Essex, Department of
  Computer Science, NLE Group, 2002.

\bibitem[Power et~al.(1999)Power, Doran, and Scott]{PowerDoranScott1999}
Richard Power, Christine Doran, and Donia Scott.
\newblock Generating embedded discourse markers from rhetorical structure.
\newblock In \emph{Proceedings of EWNLG 1999}, pages 30--38, Toulouse, 1999.

\bibitem[Prasad et~al.(2008)Prasad, Dinesh, Lee, Miltsakaki, Robaldo, Joshi,
  and Webber]{PrasadEtAl2008}
Rashmi Prasad, Nikhil Dinesh, Alan Lee, Eleni Miltsakaki, Livio Robaldo,
  Aravind Joshi, and Bonnie Webber.
\newblock The {Penn Discourse Treebank} 2.0.
\newblock In \emph{Proceedings of the 6th International Conference on Language
  Resources and Evaluation (LREC 2008)}, pages 2961--2968, Marrakesh, Morocco,
  2008.

\bibitem[Prasad et~al.(2014)Prasad, Webber, and Joshi]{PrasadWebberJoshi2014}
Rashmi Prasad, Bonnie Webber, and Aravind Joshi.
\newblock Reflections on the {Penn Discourse TreeBank}, comparable corpora, and
  complementary annotation.
\newblock \emph{Computational Linguistics}, 40\penalty0 (4):\penalty0 921--950,
  2014.

\bibitem[Prasad et~al.(2019)Prasad, Webber, Lee, and
  Joshi]{PrasadWebberLeeEtAl2019}
Rashmi Prasad, Bonnie Webber, Alan Lee, and Aravind Joshi.
\newblock {Penn Discourse Treebank Version 3.0. LDC2019T05}, 2019.

\bibitem[Roberts(2012)]{roberts:2012:information}
Craige Roberts.
\newblock Information structure in discourse: Towards an integrated formal
  theory of pragmatics.
\newblock \emph{Semantics and Pragmatics}, 5\penalty0 (6):\penalty0 1--69,
  December 2012.

\bibitem[Scheffler and Stede(2016)]{SchefflerStede2016}
Tatjana Scheffler and Manfred Stede.
\newblock Adding semantic relations to a large-coverage connective lexicon of
  {G}erman.
\newblock In \emph{Proceedings of LREC 2016}, pages 1008--1013, Reykjavik,
  Iceland, 2016.

\bibitem[Stede(2012)]{Stede2012}
Manfred Stede.
\newblock \emph{Discourse Processing}.
\newblock Synthesis Lectures on Human Language Technologies 4. Morgan \&
  Claypool, [San Rafael, CA], 2012.

\bibitem[Stede and Umbach(1998)]{StedeUmbach1998}
Manfred Stede and Carla Umbach.
\newblock {DiMLex}: A lexicon of discorse markers for text generation and
  understanding.
\newblock In \emph{Proceedings of COLING-ACL 1998}, pages 1238--1242, Montreal,
  1998.

\bibitem[Taboada(2006)]{taboada2006discourse}
Maite Taboada.
\newblock Discourse markers as signals (or not) of rhetorical relations.
\newblock \emph{Journal of pragmatics}, 38\penalty0 (4):\penalty0 567--592,
  2006.

\bibitem[Taboada and Das(2013)]{TaboadaDas2013}
Maite Taboada and Debopam Das.
\newblock Annotation upon annotation: Adding signalling information to a corpus
  of discourse relations.
\newblock \emph{Dialogue and Discourse}, 4\penalty0 (2):\penalty0 249--281,
  2013.

\bibitem[Taboada and de~los~\'{A}ngeles
  G\'{o}mez-Gonz\'{a}lez(2012)]{TaboadaAngelesGomez-Gonzalez}
Maite Taboada and Mar\'{i}a de~los~\'{A}ngeles G\'{o}mez-Gonz\'{a}lez.
\newblock Discourse markers and coherence relations: Comparison across markers,
  languages and modalities.
\newblock \emph{Linguistics and the Human Sciences}, 6\penalty0
  (1--3):\penalty0 17--41, 2012.

\bibitem[Toldova et~al.(2017)Toldova, Pisarevskaya, Ananyeva, Kobozeva,
  Nasedkin, Nikiforova, Pavlova, and
  Shelepov]{ToldovaPisarevskayaAnanyevaEtAl2017}
Svetlana Toldova, Dina Pisarevskaya, Margarita Ananyeva, Maria Kobozeva,
  Alexander Nasedkin, Sofia Nikiforova, Irina Pavlova, and Alexey Shelepov.
\newblock Rhetorical relation markers in {Russian RST Treebank}.
\newblock In \emph{Proceedings of the 6th Workshop Recent Advances in RST and
  Related Formalisms}, pages 29--33, Santiago de Compostela, Spain, 2017.

\bibitem[{Torabi Asr} and Demberg(2013)]{TorabiAsrDemberg2013}
Fatemeh {Torabi Asr} and Vera Demberg.
\newblock On the information conveyed by discourse markers.
\newblock In \emph{Proceedings of the Fourth Annual Workshop on Cognitive
  Modeling and Computational Linguistics ({CMCL})}, pages 84--93, Sofia,
  Bulgaria, 2013.

\bibitem[Versley and Gastel(2013)]{VersleyGastel2013}
Yannick Versley and Anna Gastel.
\newblock Linguistic tests for discourse relations in the {T\"{u}Ba-D/Z} corpus
  of written {G}erman.
\newblock \emph{Dialogue and Discourse}, 4\penalty0 (2):\penalty0 142--173,
  2013.

\bibitem[Webber and Joshi(1998)]{webber1998anchoring}
Bonnie~Lynn Webber and Aravind~K Joshi.
\newblock Anchoring a lexicalized {T}ree-{A}djoining {G}rammar for discourse.
\newblock In \emph{Proceedings of SIGDIAL 1998}, pages 86--92, Montreal, 1998.

\bibitem[Wiegreffe and Pinter(2019)]{WiegreffePinter2019}
Sarah Wiegreffe and Yuval Pinter.
\newblock Attention is not not explanation.
\newblock In \emph{Proceedings of EMNLP 2019}, pages 11--20, Hong Kong, China,
  2019.

\bibitem[Yu et~al.(2019)Yu, Zhu, Liu, Liu, Peng, Gong, and
  Zeldes]{YuZhuLiuEtAl2019}
Yue Yu, Yilun Zhu, Yang Liu, Yan Liu, Siyao Peng, Mackenzie Gong, and Amir
  Zeldes.
\newblock {GumDrop} at the {DISRPT2019} shared task: A model stacking approach
  to discourse unit segmentation and connective detection.
\newblock In \emph{Proceedings of Discourse Relation Treebanking and Parsing
  (DISRPT 2019)}, pages 133--143, Minneapolis, MN, 2019.

\bibitem[Zeldes(2017)]{zeldes2017gum}
Amir Zeldes.
\newblock The {GUM} {C}orpus: {C}reating {M}ultilayer {R}esources in the
  {C}lassroom.
\newblock \emph{Language Resources and Evaluation}, 51\penalty0 (3):\penalty0
  581--612, 2017.

\bibitem[Zeldes(2018{\natexlab{a}})]{Zeldes2018}
Amir Zeldes.
\newblock \emph{Multilayer Corpus Studies}.
\newblock Routledge Advances in Corpus Linguistics 22. Routledge, London,
  2018{\natexlab{a}}.

\bibitem[Zeldes(2018{\natexlab{b}})]{zeldes2018neural}
Amir Zeldes.
\newblock A neural approach to discourse relation signaling.
\newblock In \emph{Georgetown University Round Table (GURT) 2018: Approaches to
  Discourse}, Washington, DC, 2018{\natexlab{b}}.

\bibitem[Zeyrek et~al.(2013)Zeyrek, Demir\c{s}ahin, Sevdik-\c{C}all{\i}, and
  \c{C}ak{\i}c{\i}]{ZeyrekDemirsahinSevdik-CalliEtAl2013}
Deniz Zeyrek, I\c{s}in Demir\c{s}ahin, Ay{\i}\c{s}{\i}\u{g}{\i}~B.
  Sevdik-\c{C}all{\i}, and Ruket \c{C}ak{\i}c{\i}.
\newblock {Turkish Discourse Bank}: Porting a discourse annotation style to a
  morphologically rich language.
\newblock \emph{Dialogue and Discourse}, 4\penalty0 (2):\penalty0 174--184,
  2013.

\bibitem[Zhou et~al.(2014)Zhou, Lu, Zhang, and Xue]{ZhouLuZhangEtAl2014}
Yuping Zhou, Jill Lu, Jennifer Zhang, and Nianwen Xue.
\newblock {Chinese Discourse Treebank 0.5 LDC2014T21}, 2014.

\end{thebibliography}

\end{document}